\documentclass[10pt,journal,compsoc]{IEEEtran}

\ifCLASSOPTIONcompsoc
  \usepackage[nocompress]{cite}
\else
  \usepackage{cite}
\fi

\usepackage{graphicx}

\usepackage{tikz}
\usepackage{comment}
\usepackage{amsmath,amssymb} % define this before the line numbering.
\usepackage[colorlinks,linkcolor=red]{hyperref}

\usepackage{multirow}
\usepackage{algpseudocode}  
\usepackage{amsmath}  
\usepackage{array}
\usepackage[ruled,linesnumbered]{algorithm2e}
\usepackage{hyperref}
\usepackage{colortbl}

\begin{document}
\title{Distributional Modeling for\\  Location-Aware Adversarial Patches}
%Trap Regions: Exploring the Robustness of Visual Recognition to Adversarial Location
%Distributional Modeling for Adversarial Regions: Exploring the Robustness of Locational Occlusion in Face Recognition


\author{Xingxing~Wei,~
        Shouwei~Ruan,~
        Yinpeng~Dong,~
        and~Hang Su % <-this % stops a space
\IEEEcompsocitemizethanks{\IEEEcompsocthanksitem Xingxing Wei and Shouwei Ruan are with the Institute of Artificial Intelligence, Beihang University, No.37, Xueyuan Road, Haidian District, Beijing, 100191, P.R. China. (E-mail: {shouweiruan, xxwei}@buaa.edu.cn).
Yinpeng Dong and Hang Su are with the Institute for Artificial Intelligence, Beijing National Research Center for Information Science and Technology, Department of Computer Science and Technology, Tsinghua University, Beijing 100084, China.
\IEEEcompsocthanksitem Corresponding author: Xingxing Wei.}% <-this % stops an unwanted space
}


% \markboth{IEEE Transactions on Pattern Analysis and Machine Intelligence}%
% {Shell \MakeLowercase{\textit{et al.}}: Bare Demo of IEEEtran.cls for Computer Society Journals}

\IEEEtitleabstractindextext{%
\begin{abstract}
Adversarial patch is one of the important forms of performing adversarial attacks in the physical world. To improve the naturalness and aggressiveness of existing adversarial patches, location-aware patches are proposed, where the patch's location on the target object is integrated into the optimization process to perform attacks. Although it is effective, efficiently finding the optimal location for placing the patches is challenging, especially under the black-box attack settings. In this paper, we propose the Distribution-Optimized Adversarial Patch (DOPatch), a novel method that optimizes a multimodal distribution of adversarial locations instead of individual ones. DOPatch has several benefits: Firstly, we find that the locations' distributions across different models are pretty similar, and thus we can achieve efficient query-based attacks to unseen models using a distributional prior optimized on a surrogate model. Secondly, DOPatch can generate diverse adversarial samples by characterizing the distribution of adversarial locations. Thus we can improve the model's robustness to location-aware patches via carefully designed Distributional-Modeling Adversarial Training (DOP-DMAT). We evaluate DOPatch on various face recognition and image recognition tasks and demonstrate its superiority and efficiency over existing methods. We also conduct extensive ablation studies and analyses to validate the effectiveness of our method and provide insights into the distribution of adversarial locations.
\end{abstract}

}

\maketitle

\IEEEdisplaynontitleabstractindextext

% \fi
\IEEEpeerreviewmaketitle

\IEEEraisesectionheading{\section{Introduction}\label{sec:introduction}}

\IEEEPARstart {V}{isual} systems based on deep neural networks are vulnerable to adversarial examples~\cite{szegedy2013intriguing,goodfellow2014explaining}. Previous works have striven to explore the robustness of visual models under the $L_p$-norms adversarial examples~\cite{carlini2017towards,madry2017towards,kurakin2018adversarial,dong2018boosting}, which deceive the model by adding small perturbations to input samples. However, in the physical world, adversarial patch attacks constrained to a localized region are more commonly encountered. In many computer vision tasks, such as face recognition, adversarial patch attacks are often executed by objects accompanying the human's face (e.g., glasses~\cite{sharif2016accessorize, sharif2019general}, masks~\cite{zolfi2021adversarial}, hats~\cite{komkov2021advhat}, etc.) or using an optimized patch with or without semantic information~\cite{brown2017adversarial,lee2019physical,li2021generative,wei2022adversarial, wei2022simultaneously}. Such attacks are more threatening and easier to execute in the physical world,  and can also be  inconspicuous, thus posing a significant challenge to applying deep networks in security-critical applications. Therefore, evaluating and enhancing the robustness of visual models against adversarial patches is essential.

\begin{figure}[t]
  \centering
  \includegraphics[width=0.99\linewidth]{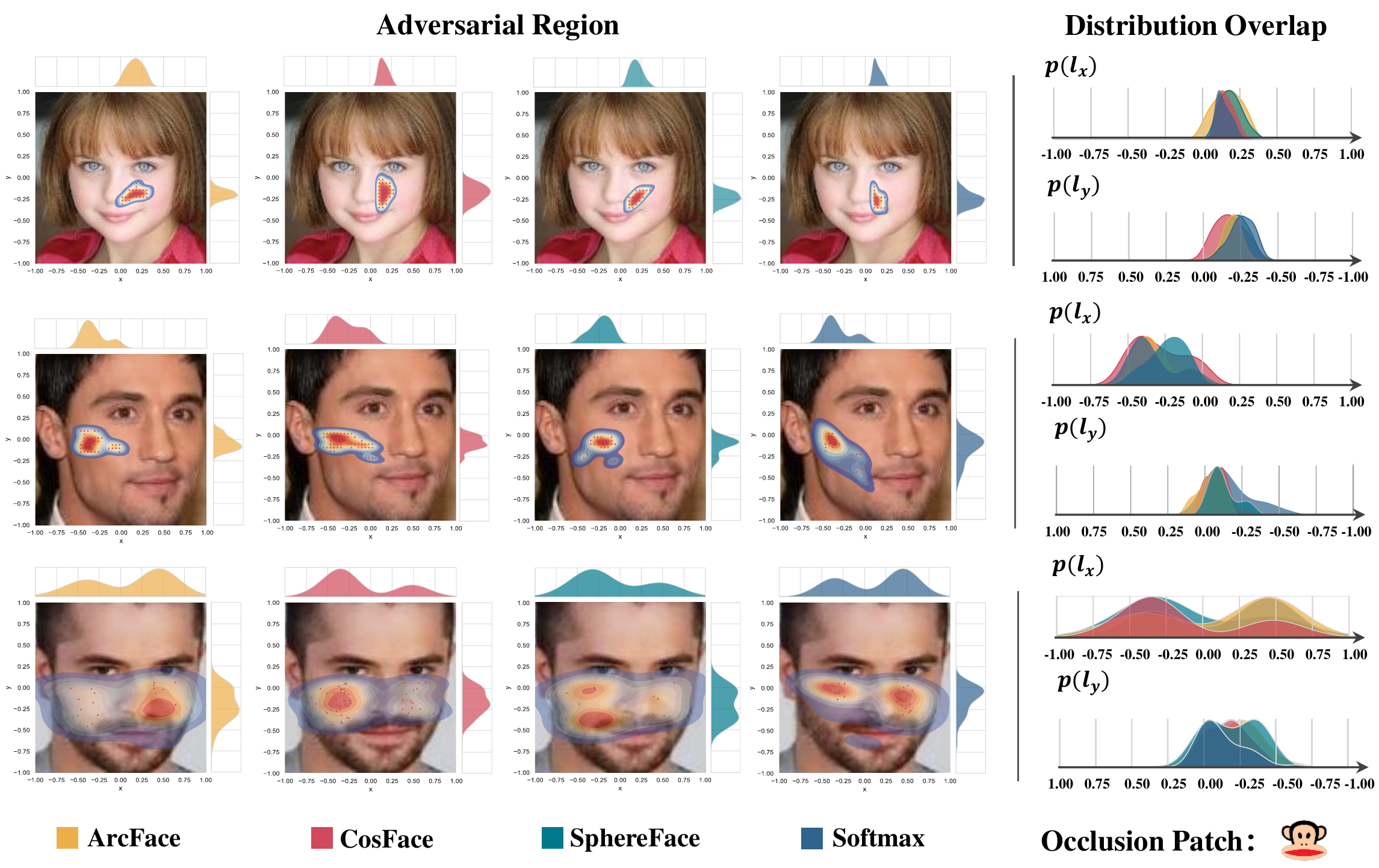}
   \caption{When the patch is applied on the faces, the adversarial locations are regional aggregation (left) and the distributions are pretty similar across  models (right). To ensure passing live detection, we mask the regions of  facial features when optimizing  adversarial locations like ~\cite{wei2022adversarial}.}
   \label{fig:0}\vspace{-0.5cm}
\end{figure}

Generating adversarial patches involves optimizing unrestricted perturbations within localized regions. Traditional adversarial patches usually aim at optimizing perturbations while fixing the position of the object. Although regularization measures, such as Total Variance (TV) loss and Non-Printability Score (NPS) loss~\cite{komkov2021advhat,sharif2019general}, are incorporated to enhance the robustness and smoothness of adversarial perturbations, these attacks remain less natural, easily detectable, and defensible due to the local semantic deficit caused by the unrestricted perturbations~\cite{chiang2020certified, xiang2021patchguard,rao2020adversarial}. To achieve both natural and aggressive patches, recent studies investigate the impact of the patch's location~\cite{wei2022adversarial,wei2022simultaneously,rao2020adversarial,li2021generative}, \cite{yang2020patchattack}, which is called location-aware adversarial patches in our paper. Location-aware patches are flexible, they can not only combine with the fixed pattern to construct the natural and stealthy physical attack (e.g., adversarial sticker in~\cite{wei2022adversarial}) but also conduct the joint optimization with the traditional patches to improve their attack performance (e.g., \cite{wei2022simultaneously, li2021generative}, etc.). For these reasons, location-aware patches deserve further investigation. 

%In more extreme cases, face recognition models can be effectively attacked by fixed patterns (e.g., stickers~\cite{wei2022adversarial, guo2021meaningful}) with an optimized placement location (2D coordinates and rotate angle), which is called location-optimized patches. Furthermore, a recent observation reveals that adversarial locations of fixed patterns exhibit significant regional aggregation, which strongly motivates our research in this paper.

\begin{figure*}[thp]
  \centering
  \includegraphics[width=0.99\linewidth]{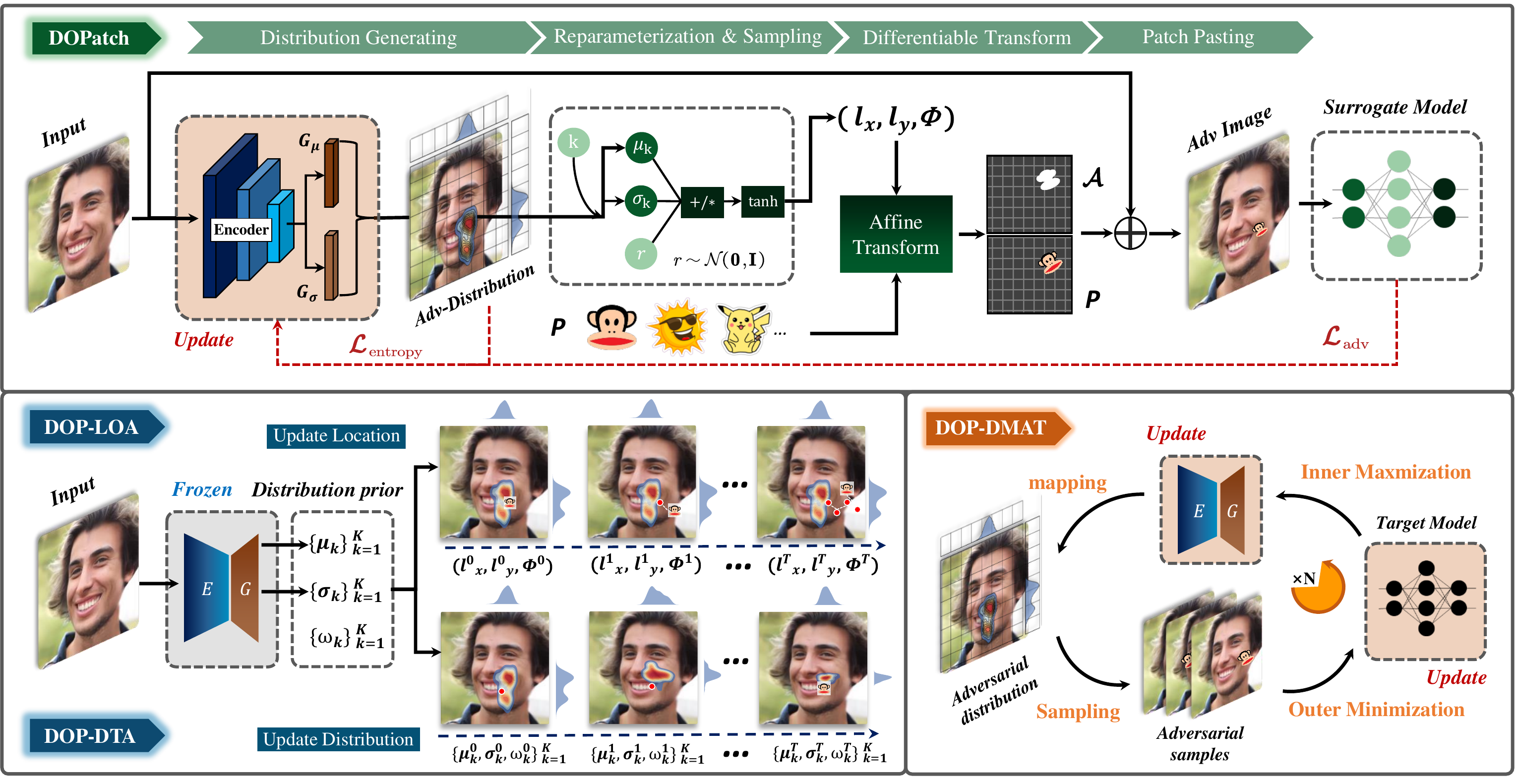}
   \caption{The overview of our method. The first stage (DOPatch) is carried out on the surrogate model to optimize a distribution mapping network. It maps the image input to the corresponding distribution parameters of the adversarial location. The second stage is to conduct attacks (DOP-LOA and DOP-DTA) or defense (DOP-DMAT) leveraging the distributional prior. To better show the effect of the distribution-optimized modeling method in our paper, we adopt the same setting in~\cite{wei2022adversarial} and only optimize the patch's location for the fixed meaningful pattern. Thus, we can focus on location optimization and don't need to consider the pattern's influence.}
   \label{fig:framework}\vspace{-0.5cm}
\end{figure*}

Despite the vulnerability of models to such location-aware patch attacks, current methods have certain limitations. For example, a recent observation~\cite{wei2022adversarial} reveals that adversarial locations of fixed patterns exhibit significant regional aggregation, as shown in Fig.~\ref{fig:0}, where the colorful regions on the faces denote the adversarial locations to show effective attacks. We see that the positions that can lead to successful attacks are not discretely distributed but clustered in some specific regions. Due to this phenomenon, the main limitation is that existing methods concentrate solely on optimizing a single location~\cite {wei2022adversarial,li2021generative,wei2022simultaneously} and neglect their intrinsic distribution property, thus leading to inadequate modeling to characterize the aggregation regions of the adversarial locations, and further restricting its potential function on attack efficiency and defense effects.

%which is (1) challenging to optimize for the worst-case sample using gradient-based methods (as proved in~\cite{engstrom2019exploring,venkatesh2023adversarial}), (2) difficult to perform effective transfer attacks across models and (3) inadequate to characterize the aggregation regions of the adversarial locations. Furthermore, since optimising adversarial locations primarily relies on black-box-based methods~\cite{rao2020adversarial,wei2022adversarial,wei2022simultaneously}, generating diverse adversarial samples is time-consuming. Thus, improving the model's robustness to location-optimized patches via adversarial training is intractable.

To address these problems, in this paper, we propose \textbf{Distribution-Optimized Adversarial Patch (DOPatch)}, a novel method to search for the worst-case placement region in a distributional modeling manner. Specifically, DOPatch optimizes a multimodal distribution of adversarial locations instead of individual ones to characterize the multiple regions, which can provide several benefits. Firstly, we can achieve efficient query-based attacks on other unseen models using the distributional prior optimized on the surrogate model. It is based on our finding that although individual adversarial locations vary across models, there is a similarity in their distributions, as shown in Fig.~\ref{fig:0}. Secondly, DOPatch can generate adversarial samples with improved diversity by characterizing the distribution of adversarial locations. Thus, we can improve the model's robustness to location-aware patches via adversarial training. 

However, distributional modeling on the adversarial locations is challenging. A straightforward way is to exhaustively search for all the adversarial locations and then obtain the multimodal distribution via a statistical approach, which is high-computation and time-consuming. For that,  we propose a novel distribution mapping network, which maps the face images to their corresponding multimodal  distribution parameters. The network is trained under the supervision of the designed adversarial loss (Eq.(\ref{eq:L_adv_1})) and entropy regularization loss (Eq.\ref{eq:L_entropy_2}), aiming to balance the aggressiveness and generalization of the adversarial distribution. To ensure gradient propagation throughout the pipeline, we employ differentiable affine transformations to map fixed-pattern patches and their masks to the specified locations in the object by considering the object's surface change. The process is illustrated in Fig.~\ref{fig:framework} (top row).
%Technically, our framework is divided into two stages, as illustrated in Fig.~\ref{fig:framework}. The first stage aims to optimize the distribution prior to the adversarial location in the surrogate model, i.e., DOPatch. To optimize the distribution prior more effectively on the dataset, 

% To address discontinuous loss landscapes~\cite{engstrom2019exploring}, we perform multiple Monte Carlo samplings within the distribution to obtain a smoothing gradient and apply entropy regularization to control the smoothing level. 

After distributional modeling, we conduct the attack or defense leveraging the distributional transfer prior. Specifically, as for attack, we perform a query-based attack with few queries to target models, employing two distinct modes tailored to different mechanisms. The first one is \textbf{Location-Optimization Attack (DOP-LOA)}, which optimizes a single adversarial location starting from the center of the distributional prior. And the second one is  \textbf{Distribution-Transfer Attack (DOP-DTA)}, which optimizes the distribution parameters to capture the adversarial region of the black-box target model precisely. Compared with DOP-LOA, DOP-DTA makes full use of the intrinsic distribution property of adversarial locations, and thus has better attack performance. The process is illustrated in Fig.~\ref{fig:framework} (bottom left). 

As for defense, we utilize adversarial training to improve the model's robustness to the existing location-aware patches. For that, we design a \textbf{Distributional-Modeling Adversarial Training} (\textbf{DOP-DMAT}), which is formulated as a distribution-based min-max problem following the ADT~\cite{dong2020adversarial}, where the inner maximization aims to optimize the distribution of adversarial location using DOPatch. And outer minimization aims to train a robustness model by minimizing the expected loss over the worst-case adversarial location distributions. By exploring the diverse adversarial locations sampling from the distribution, DOP-DMAT is suitable for training robust models. The process is illustrated in Fig.~\ref{fig:framework} (bottom right).

\begin{figure*}[t]
  \centering
  \includegraphics[width=0.99\linewidth]{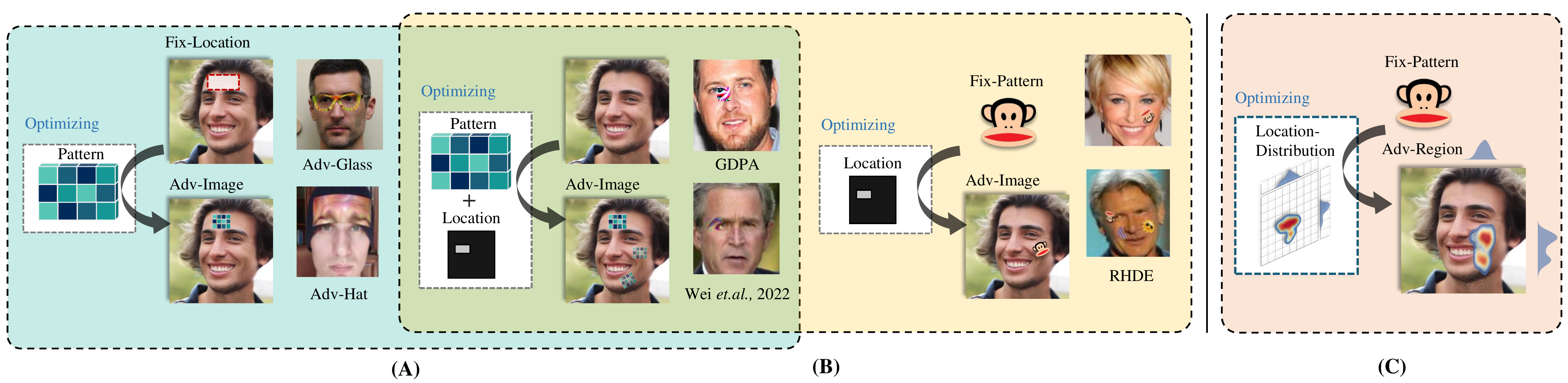}
   \caption{The summary of the existing adversarial patch attacks in face recognition: previous works can be divided into (A) \textbf{pattern-aware patches} and (B) \textbf{location-aware patches}. In this paper, we propose a new approach which optimizes the distribution of adversarial locations to improve the performance of location-aware patches, as shown in (C).}
   \label{fig:3} 
\end{figure*}

We conduct extensive experiments on the face recognition task and image recognition task to demonstrate the effectiveness of the proposed method. Experimental result shows that DOPatch characterizes the region of adversarial location and improves the attack's effectiveness while reducing the number of queries, demonstrating better performance than previous methods. Based on it, DOP-DMAT significantly improves the robustness of face recognition models against location-aware patches. The code is available at \url{https://github.com/Heathcliff-saku/DOPatch}. 

In summary, this paper has the following contributions:

\textbf{(1)} We propose the Distribution-Optimized Adversarial Patch (DOPatch), a novel framework that can efficiently search for the worst-case region of fixed-pattern patch locations. It aims to optimize a multimodal distribution of adversarial locations instead of individual location points. 

\textbf{(2)} To evaluate the robustness of black-box visual systems with limited information. We further design two query-based attack modes using distribution prior: Location Optimization Attack (DOP-LOA) and Distribution Transfer Attack (DOP-DTA), which can perform effective attacks with only a few queries with the visual systems.

\textbf{(3)} We propose DOP-DMAT, a distribution-based adversarial training that employs DOPatch to address the inner maximization problem. By leveraging DOP-DMAT, we can significantly improve the robustness of the model against location-aware adversarial patches.

\textbf{(4)}  We conduct extensive experiments on face recognition and image recognition tasks to verify the proposed methods. Specifically, we adopt DOPatch to perform attacks and defense on various face and image recognition models (including the popular big models). A variety of experiments show that our method achieves the best performance compared with the SOTA methods.

\section{Related Work}
%This section details the existing adversarial patch attacks and corresponding defense methods.

\subsection{Adversarial Patch Attacks}
\label{sec:Adversarial Patch Attacks}
The adversarial patch is originally proposed in~\cite{brown2017adversarial} to fool deep networks in the physical world. It aims to optimize an unrestricted perturbation within a particular region of a clean image. Unlike imperceptible perturbations based on $L_p$-norms, adversarial patches are capturable and robust perturbations that can be applied effectively in physical attacks. The adversarial patches are widely adopted in various vision tasks, such as image recognition~\cite{brown2017adversarial,eykholt2018robust,karmon2018lavan,liu2019perceptual}, object detection~\cite{liu2018dpatch,lee2019physical,chen2021camdar,thys2019fooling}, semantic segmentation~\cite{mirsky2021ipatch,nesti2022evaluating}, depth estimation~\cite{yamanaka2020adversarial}, etc., posing a significant threat to the deployment of models in the physical world.  Previous works propose several strategies to craft more robust patches, among which Expectation over Transformation (EOT)~\cite{athalye2018synthesizing}  is a commonly used method to improve the robustness of patches against rotation and scaling. To make adversarial patches more imperceptible,  Chindaudom et al.~\cite{chindaudom2020adversarialqr} propose the adversarial QR patch, which constrains patches to QR patterns. Additionally, Liu et al.~\cite{liu2019perceptual} propose PS-GAN, the first GAN-based method to generate more natural-looking adversarial patches. Adversarial patches also threaten face recognition tasks, where the imperceptibility of patches is a more significant concern than other tasks, like Adv-Glasses~\cite{sharif2016accessorize,sharif2019general}, Adv-Hat~\cite{komkov2021advhat}, Adv-masks~\cite{zolfi2021adversarial}, AT3D~\cite{yang2023towards}, etc. These method mainly depend on optimizing the patch's pattern to achieve attacks, thus we can call them as \textbf{pattern-aware adversarial patches}. 

On the contrary, some studies aim at optimizing the patch's location by using the fixed meaningful pattern to perform attacks. For example,  RHDE~\cite{wei2022adversarial} achieves the attack by optimizing the placement of an existing sticker. Not only obtaining the inferior in attack performance, it is also easy to implement, i.e., it can use common materials such as cartoon stickers as the fixed pattern. Inspired by these works, the joint pattern-location optimization methods are proposed. For example, Li et al.~\cite{li2021generative}  generate the texture and location of patches using a generator network. Similarly, Wei et al.~\cite{wei2022simultaneously} propose a reinforcement learning framework for joint optimization of texture and location in a black-box setting. \cite{rao2020adversarial} performs pattern-location optimization via an iterative strategy. %, and Patchattack \cite{yang2020patchattack} introduces a black-box texture-based attack with changeable locations. 
Because the patch's location is integrated to conduct attacks, we  call them as \textbf{location-aware adversarial patches}.

The summary for these related works is given in Fig.~\ref{fig:3}. While the location-aware adversarial patches show good performance, they are all limited to optimizing individual positions. In contrast, we aim to model the distribution of adversarial locations implicitly. This allows us to comprehensively characterize these regions and improve the performance and robustness of location-aware patches. Moreover, we can efficiently improve the model's robustness against the patch through distribution-based adversarial training. To better show the effect of the distribution-optimized modeling method in our paper, we adopt the same setting in~\cite{wei2022adversarial} and only optimize the patch's location for the fixed pattern. Thus, we can focus on the location optimization, and don't need to consider the pattern's influence.

\subsection{Defense for Adversarial Patches}
\label{sec:Spatial Robustness in Vision Recognition}

Various methods have been proposed to defend against adversarial patches, which can be categorized into three stages: pre-processing, in-processing, and post-processing. Pre-processing methods apply some transformations to the input data to increase the model’s robustness. For example, removing \cite{hayes2018visible, liu2022segment} or blurring \cite{naseer2019local} the patch in input images can mitigate the effect of adversarial attacks. In-processing methods use some techniques to reduce or avoid the impact of adversarial attacks during the model’s training phase, such as adversarial training \cite{rao2020adversarial, metzen2021meta}.  Post-processing methods process the model’s output to correct the wrong prediction. For instance, \cite{gurel2021knowledge} adjusts the predicted probabilities to a more balanced distribution to ensure correct prediction.

Our method belongs to the in-processing stage. But different the existing methods \cite{ rao2020adversarial, metzen2021meta} that utilize pattern-aware adversarial patches to perform adversarial training, we use location-aware adversarial patches, and furthermore conduct adversarial distribution training, which can achieve better defense performance.

\section{Methodology}
This section first introduces the distributional transferability of adversarial locations in Sec.~\ref{sec:distributional transferability}.  Based on this finding, we formulate the problem in Sec.~\ref{sec:Problem Formulation} and present the distributional modeling and black-box attacks in Sec.~\ref{sec:Distribution-Optimized Adversarial Patch} and Sec.~\ref{sec:Black-Box Attack with Distributional Prior}. Finally, we give the adversarial training method in Sec.~\ref{sec:Distribution-Based Adversarial Training}.  An overview of our framework is shown in Fig. \ref{fig:framework}.

\subsection{Distributional Transferability}
\label{sec:distributional transferability}
Previous work finds that adversarial locations have regional aggregation when a fixed patch pattern’s perturbations is given \cite{wei2022adversarial,wei2022simultaneously}. We further investigate the distribution of these aggregate locations for the same sample across different models and discover an interesting phenomenon: \textbf{the distributions of adversarial locations on different models partially overlap, indicating that there exists a cross-model transferability in the distribution of adversarial locations.}

To verify this hypothesis, we first devise a preliminary experiment to observe the distribution of adversarial locations and explore their transferability. Specifically, the face classification task serves as an example, where we begin by grid-traversing each location in an image sample ${\mathbf{X}}$ to capture the corresponding adversarial locations $\boldsymbol{\theta}\!=\!(l_x, l_y)$ causing successful attacks. Next, we use kernel density estimation (KDE) to carve the joint probability density $p(\boldsymbol{\theta})$ for the given ${\mathbf{X}}$, w.r.t various face classifiers that are trained with different loss functions, including ArcFace~\cite{deng2019arcface}, CosFace~\cite{wang2018cosface}, SphereFace~\cite{liu2017sphereface} and Softmax~\cite{taigman2014deepface,sun2014deep}. The visualization for a face image is shown in Fig.~\ref{fig:0} (left), where we observe that the joint distributions $p(\boldsymbol{\theta})$ across different models exhibit high similarity. To illustrate this point more visually, we also depict the marginal distributions $p(l_x)$, $p(l_y)$ in Fig.~\ref{fig:0} (right), highlighting their significant overlap. We conduct experiments on 100 face images randomly selected from LFW dataset, and the similar phenomenon occurs on all the samples. This observation inspires the idea that if we can optimize the distribution of adversarial locations in surrogate models, we can leverage its cross-model transfer prior to perform efficient query-based black-box attacks \cite{andriushchenko2020square}. To this end, we propose two methods to optimize the distribution of adversarial locations, which are detailed in Sec. \ref{sec:Black-Box Attack with Distributional Prior}. Besides, because $p(\boldsymbol{\theta})$ captures the diversity of adversarial locations, we can perform an effective adversarial training using the diverse adversarial locations sampling from the distribution, details are given in Sec. \ref{sec:Distribution-Based Adversarial Training}.

\subsection{Problem Formulation}
\label{sec:Problem Formulation}
In visual recognition,  the adversarial attack aims to apply a perturbation to the clean image $\mathbf{X}$ to make the model predict incorrectly. For the $L_p$-norms attack, the perturbation $\boldsymbol{\delta}$ is added to each pixel over the entire image, formulated as $\mathbf{X}^{adv}=\mathbf{X}+\boldsymbol{\delta}$. In contrast, the perturbation of the adversarial patch is limited to a particular shape and location:
\begin{equation}
\mathbf{X}^{\mathrm{adv}}=(1-\boldsymbol{\mathcal{A}})\odot\mathbf{X}+\boldsymbol{\mathcal{A}}\odot\boldsymbol{\delta},
\label{eq:patch attack}
\end{equation}
where $\boldsymbol{\mathcal{A}}\in\{0,1\}^{w\times h}$ is a binary matrix to mask the patch area and $\odot$ represents the Hadamard product. The pattern-aware methods focus on optimizing the patch perturbation $\boldsymbol{\delta}$ with a pre-fixed $\boldsymbol{\mathcal{A}}$. In contrast, the location-aware methods do not generate $\boldsymbol{\delta}$ but choose a fixed pattern as perturbations, such as cartoon stickers~\cite{wei2022adversarial}, and fix the shape of $\boldsymbol{\mathcal{A}}$, while trying to optimize the pasting location to generate adversarial patch samples as follows:
\begin{equation}
\mathbf{X}^{\mathrm{adv}}=(1-T(\boldsymbol{\mathcal{A}},\boldsymbol{\theta}))\odot\mathbf{X}+T(\boldsymbol{\mathcal{A}},\boldsymbol{\theta})\odot\mathbf{P},
\label{eq:location patch attack}
\end{equation}
where $\mathbf{P}$ is the fixed pattern, $T(\cdot)$ is the affine transform operator, and $\boldsymbol{\theta}=[l_x,l_y,\phi]$ represents the affine transformation parameters, where $l_x$ and $l_y$ are the pattern translation along the x and y axes, and $\phi$ is the pattern rotation. For simplicity, we define $\mathcal{G}(\mathbf{X},\mathbf{P},\boldsymbol{\theta})$ as the process of generating adversarial samples using the location-aware patch.

Unlike previous methods searching for a single adversarial location $\boldsymbol{\theta}$, the proposed DOPatch captures the aggregated regions of the adversarial locations using a distribution $p(\boldsymbol{\theta})$. By optimizing the $p(\boldsymbol{\theta})$ under proper parameterization, most $\boldsymbol{\theta}$ sampled from it is likely to mislead the model, and we can use $p(\boldsymbol{\theta})$ for effective query-based attacks. Thus, based on Eq.~\eqref{eq:location patch attack}, we formulate DOPatch as solving the following optimization problem:
\begin{equation}
\max_{p(\boldsymbol{\theta})} \left \{ \mathbb{E}_{p(\boldsymbol{\theta})}\left [ \mathcal{L}\left (\mathcal{G}(\mathbf{X},\mathbf{P},\boldsymbol{\theta}) \right ) - \lambda\cdot \log p(\boldsymbol{\theta}) \right ]  \right \},
\label{eq:max problem}
\end{equation}
where $\mathcal{L}(\cdot)$ is the adversarial loss (\emph{e.g.} cross-entropy loss in classification task) that will be specifically defined in Sec.~\ref{sec:Distribution-Optimized Adversarial Patch}, $-\log p(\boldsymbol{\theta})$ is the negative log density, whose expectation is the distribution’s entropy, we use it as a regularization loss to avoid the optimal distribution degenerate into a Dirac one~\cite{dong2020adversarial}. $\lambda$ is the balance parameter between adversarial and entropy losses.

As shown in Eq.~\eqref{eq:max problem}, DOPatch aims to optimize the worst-case distribution of adversarial locations by maximizing the loss expectation. The motivation for using $p(\boldsymbol{\theta})$ instead of individual $\boldsymbol{\theta}$ is multifaceted. \textbf{\emph{Firstly}}, as explored in Sec.~\ref{sec:distributional transferability}, using distribution prior can carry out efficient query-based attacks due to its cross-model transferability prior. \textbf{\emph{Secondly}}, learning the underlying distribution to explore the adversarial location region can better help us understand the model's vulnerabilities. %\textbf{\emph{Thirdly}}, due to the discontinuity of spatial transformations in the loss landscape, previous methods have difficulty in first-order optimization of spatial parameters~\cite{engstrom2019exploring}. In contrast, our method aims to maximize the expected loss within the distribution, which can achieve optimization on a smoother loss landscape.
% 定义人脸识别攻击，定义patch攻击，以及位置参数。location optim patch的最大化目标
% 改为使用分布的最大化目标 并给出对抗性训练的min-max形式

\subsection{Distribution-Optimized Adversarial Patch}
\label{sec:Distribution-Optimized Adversarial Patch}

\subsubsection{Parameterization of Location Distributions}

A natural approach to solving Eq.~\eqref{eq:max problem} is to use trainable parameters to parameterize the adversarial distribution $p(\boldsymbol{\theta})$. Since the loss landscape w.r.t adversarial locations exhibits multiple local maximal regions, we expect the optimized $p(\boldsymbol{\theta})$ to be able to capture the adversarial locations comprehensively. Previous works, such as Adversarial Distributional Training (ADT)~\cite{dong2020adversarial}, use the uni-modal Gaussian distribution to parameterize the adversarial distribution of the $L_p$-norms perturbation. However, the uni-modal Gaussian distribution is insufficient to characterize multiple local maximals. To overcome this problem, we use multimodal Gaussian distribution as the parametric representation of $p(\boldsymbol{\theta})$, which learns multiple Gaussian components to cover different local maximals of the loss landscape.

Specifically, we parameterize $p(\boldsymbol{\theta})$ by a mixture of K Gaussian components, in which the distribution parameters are denoted as $\Psi=\{\boldsymbol{\mu}_k, \boldsymbol{\sigma}_k, \omega_k\}_{k=1}^K$. The $\boldsymbol{\mu}_k\in\mathbb{R}^3$ and $\boldsymbol{\sigma}_k\in\mathbb{R}^3$ are the mean and standard deviation of k-th Gaussian component. The $\omega_k\in [0,1](\Sigma_{k=1}^K\omega_k=1)$ represents the weights of components. To ensure that $\boldsymbol{\theta}$ is bounded within the range $[-1, 1]$ allowed by the affine transformation matrix, we further employ the transformation of the random variable approach. Eventually, $p(\boldsymbol{\theta})$ is parameterized as:
\begin{equation}\label{eq:parameterized}
 \boldsymbol{\theta} = \tanh(\mathbf{u}/\tau),\; \mathrm{with} \; p(\mathbf{u}|\Psi) = \sum_{k=1}^{K} \mathbf{\omega}_k\cdot\mathcal{N} (\mathbf{u}|\boldsymbol{\mu}_k,\boldsymbol{\sigma}^2_k\mathbf{I}),   
\end{equation}
where $\tau$ is a hyperparameter controlling the slope of $\tanh$. As shown in Eq.~\eqref{eq:parameterized},  location $\boldsymbol{\theta}$ is obtained by $\tanh$ transformation of intermediate variables $\mathbf{u}$ for normalization, while $\mathbf{u}$ follows a multi-modal Gaussian distribution. Note that $p(\mathbf{u}|\Psi)$ is now in a
summation form, which is not convenient for computing the gradient. Therefore, we rewrite $p(\mathbf{u}|\Psi)$ into a multiplication form:
\begin{equation}\label{eq:parameterized_2}
\begin{split}
p(\mathbf{u}|\Psi)&={\textstyle \sum_{\boldsymbol{\Gamma}}^{}} p(\mathbf{u},\boldsymbol{\Gamma}|\Psi),\\ \mathrm{where} \; p(\mathbf{u},\boldsymbol{\Gamma}|\Psi) &= \prod_{k=1}^{K} \mathbf{\omega}_k^{\gamma_k}\cdot\mathcal{N} (\mathbf{u}|\boldsymbol{\mu}_k,\boldsymbol{\sigma}^2_k\mathbf{I})^{\gamma_k},   
\end{split}
\end{equation}
where $\boldsymbol{\Gamma}=\{\gamma_k\}_{k=1}^K$ is a latent one-hot vector introduced to determine which component the sampled location belongs to and obeys a multinomial distribution
with probability $\omega_k$. Given the $p(\mathbf{u}|\Psi)$ parameterized by Eq.~\eqref{eq:parameterized_2}, the optimization problem in Eq.~\eqref{eq:max problem} is appropriately rewritten as:
\begin{equation}
\begin{split}
\max_{\Psi} \bigl\{ \mathbb{E}_{p(\mathbf{u},\boldsymbol{\Gamma}|\Psi)}\bigl [ &\mathcal{L}\left (\mathcal{G}(\mathbf{X},\mathbf{P},\tanh(\mathbf{u}/\tau) \right ) \\ 
& - \lambda\cdot \log p(\tanh(\mathbf{u}/\tau)) \bigr] \bigr\}.
\label{eq:max problem_2}
\end{split}
\end{equation}
To solve Eq.~\eqref{eq:max problem_2}, we adopt gradient-based methods to obtain optimized distribution parameters $\Psi$, which will be given in the next section.

\subsubsection{Distribution Mapping Network}

While it is possible to directly optimize $\Psi$ using gradient descent directly, in order to achieve greater attack efficiency and generalize the attack performance to unseen samples, we draw inspiration from prior work~\cite{li2021generative,liu2019perceptual} and propose a novel distribution mapping network, which aims to learn the mapping functions from clean samples to their corresponding adversarial location distribution parameters: 
\begin{equation}
F_{\mathbf{W}}= \mathbf{X} \to \{\boldsymbol{\mu}_k, \boldsymbol{\sigma}_k\}_{k=1}^K,
\end{equation}
where $\mathbf{W}$ is the weight of the network. This method allows for the efficient optimization of $\Psi$ in higher-dimensional parameter space and is capable of generalizing to unseen samples due to the learned mapping functions. Note that here we do not learn the component weights $\omega_k$. The reason is two-fold. First, in practice, the multi-modal distribution tends to degrade to the uni-modal case when $\omega_k$ is optimizing, thus failing to capture heterogeneous adversarial locations. Second, since  $\omega_k$ will be used to sample latent variables $\Gamma$ from multinomial distribution, this operation is non-differentiable and cannot be gradient-passed. Thus, we fix $\omega_k\!\!=\!\!1/K$ and properly optimize them in the second stage. 

We design the network based on an encoder-decoder architecture, where the encoder $E$ extracts feature from the input image $\mathbf{X}$, the decoders $G_{\mu}$, $G_{\sigma}$ share the same latent features extracted by $E$ and generate the corresponding means $\{\boldsymbol{\mu}_k\}_{k=1}^K$ and standard deviations $\{\boldsymbol{\sigma}_k\}_{k=1}^K$. Specifically, we use two different encoder architectures to cope with different target models: ResNet-50~\cite{he2016deep} and the architecture adapted from CycleGAN~\cite{zhu2017unpaired}. On the top of $E$, we use two fully-connected layers as the $G_{\mu}$ and $G_{\sigma}$, respectively. After obtaining the distribution parameters by the distribution mapping network, we adopt the low-variance reparameterization trick~\cite{blundell2015weight,kingma2013auto} to maintain the gradient information during follow-up sampling. Specifically, we reparameterize $\mathbf{u}$ as 
$\mathbf{u} = {\prod_{k=1}^{K}}\boldsymbol{\mu}^{\gamma_k}_k+{ \prod_{k=1}^{K}}\boldsymbol{\sigma}^{\gamma_k}_k\cdot\mathbf{r}$, where $\mathbf{r}\sim \mathcal{N}(\mathbf{0},\mathbf{I})$. Thus, in the classification task, the first term in Eq.~\eqref{eq:max problem_2} (i.e., the adversarial loss)  can be calculated as:
\begin{equation}
\begin{split}
\mathcal{L}_\text{adv} = \mathcal{L}_\text{CE}\bigl ( f \bigl (\mathcal{G}(\mathbf{X},\mathbf{P},\tanh({\prod_{k=1}^{K}}\frac{\boldsymbol{\mu}^{\gamma_k}_k}{\tau}+{ \prod_{k=1}^{K}}\frac{\boldsymbol{\sigma}^{\gamma_k}_k\cdot\mathbf{r}}{\tau} )\bigr ),t \bigr ),
\label{eq:L_adv_1}
\end{split}
\end{equation}
where $f$ is the victim model, $\mathcal{L}_\text{CE}$ is the cross-entropy loss, and $t$ is the true label of the input image. Similarly, in the identification task, we define the adversarial loss as:
\begin{equation}
\small
\begin{split}
\mathcal{L}_\text{adv}\! =\! 1-{\cos}\bigl ( f \bigl (\mathcal{G}(\mathbf{X},\mathbf{P},\tanh({\prod_{k=1}^{K}}\frac{\boldsymbol{\mu}^{\gamma_k}_k}{\tau}+{ \prod_{k=1}^{K}}\frac{\mathbf{\boldsymbol{\sigma}^{\gamma_k}_k\cdot \mathbf{r}}}{\tau} )\bigr ),f(\mathbf{X}) \bigr ),
\label{eq:L_adv_2}
\end{split}
\end{equation}
where $\cos(\cdot)$ is the cosine similarity. The second term in Eq.~\eqref{eq:max problem_2} (i.e., the entropy loss) can be calculated analytically as (proof in Appendix B.7):
\begin{equation}
\begin{split}
\mathcal{L}_\text{entropy}=& \sum_{k=1}^{K}\gamma_k\cdot\Big[-\boldsymbol{\omega}_k+ \frac{\mathbf{r}^2}{2}+\frac{\log(2\pi)}{2}+\log \boldsymbol{\sigma}_k \\ & +\log(\tanh(\frac{\boldsymbol{\mu}_k+\boldsymbol{\sigma}_k \mathbf{r}}{\tau})^2 -1) - \log\tau\Big].
\label{eq:L_entropy_2}
\end{split}
\end{equation}

By integrating two loss terms, the total objective of the distribution mapping network is:
\begin{equation}
\max_{\mathbf{W}} \left [ \mathbb{E}_{\mathcal{N}(\mathbf{r}|\mathbf{0},\mathbf{I}))}\left (\mathcal{L}_\text{adv} - \lambda\cdot \mathcal{L}_\text{entropy} \right )  \right ],
\label{eq:max problem_3}
\end{equation}
In practice, we use the Monte Carlo method to approximate the expectation. %Due to space limitations, training details will be given in the Appendix.

\subsubsection{Differentiable Affine Transformation}
The affine transformation is an essential step in generating location-aware adversarial patches, which involves transforming the fixed pattern to a specific location in the object by considering the object's surface change. To achieve this, we define differentiable affine transform operators based on the approach presented in Spatial Transformer Networks (STN)\cite{jaderberg2015spatial}. In contrast to GDPA\cite{li2021generative}, which considers only translation, we incorporate rotation to expand the search space. Specifically, for a given location parameter $\boldsymbol{\theta}=[l_x, l_y, \phi]\in[-1, 1]^3$ (derived from Eq.~\eqref{eq:parameterized}), we define the pixel index relation of the pattern before and after the affine transformation as:
\begin{equation}
\begin{Bmatrix} x_i^s \\ y_i^s\end{Bmatrix} = \begin{bmatrix} \cos(\frac{\phi\cdot \pi}{2}) & \sin(\frac{\phi\cdot \pi}{2}) & \frac{w\cdot l_x}{2}  \\  -\sin(\frac{\phi\cdot \pi}{2}) &  \cos(\frac{\phi\cdot \pi}{2}) & \frac{h\cdot l_y}{2}\end{bmatrix} \cdot \begin{Bmatrix} x_i^t \\ y_i^t \\ 1\end{Bmatrix},
\label{affine_1}
\end{equation}
where $(x_i^s, y_i^s)$ is the pixel index of the original image while $(x_i^t, y_i^t)$ is the corresponding index of the transformed image. To achieve a differentiable transformation, STN uses the bilinear interpolation method to sample the pixel values of the transformed image from the original image:
\begin{equation}
V_i^c\!=\!\sum_{h=1}^{H} \sum_{w=1}^{W} U_{hw}^c\max(0,1-\left | x_i^s-w \right | )\max(0,1-\left | y_i^s-h \right | ),
\label{affine_2}
\end{equation}
where $c$ denotes the colour channels, $U_{hw}^c$ is the pixel value of the original image in the index $(h,w)$, and $V_i^c$ is the pixel value of the transformed image in the index $(x_i^t, y_i^t)$. Thus, we can use differentiable affine transformations to simultaneously transform the fixed pattern and its corresponding mask, allowing the entire pipeline to backpropagate the loss.

\subsubsection{Overview of Optimization Algorithms}
As demonstrated in Fig.~\ref{fig:framework}, the pipeline of our method is divided into two stages. The core of the first stage is to learn the distribution mapping network in a surrogate white-box model to obtain the distribution prior. Formally, given a dataset $\mathcal{D}=\{(\mathbf{X}_i, t_i)\}_{i=1}^n$, for each training epoch, we first obtain the distribution parameters corresponding to each image using the distribution mapping network. Then we sample a set of location parameters for each image using the Monte Carlo method, generate a set of adversarial samples by Eq.~\eqref{eq:location patch attack}, feed them to the surrogate model, calculate the mean of losses, and backpropagate to update the distribution mapping network weights. Algorithm~\ref{al:dopatch} outlines the optimization algorithm of DOPatch in the first stage.

\begin{algorithm}
\footnotesize
\caption{\footnotesize Optimization Algorithms for DOPatch}\label{al:dopatch}
\KwIn{attack data $\mathcal{D}$, distribution mapping network $F_\mathbf{W}$, surrogate model $f$, training epochs $N$, components number $K$, Monte Carlo sample number $Q$, balance hyperparameter $\lambda$.}
Initialize network weights $\mathbf{W}$\;
$\{\omega_k\}_{k=1}^K \leftarrow \frac{1}{K}$\;
\For{$\mathrm{epoch}=1\;\mathbf{to}\; N$}{
\For{$\mathrm{each\;minbatch} \;\mathcal{B}\subset\mathcal{D}$}{
\tcc{obtaining distribution parameters}
$\{\boldsymbol{\mu}_k^*, \boldsymbol{\sigma}_k^*\}_{k=1}^K \leftarrow F_\mathbf{W}(\mathcal{B})$ \;
$\mathcal{L} \leftarrow 0$\;
\tcc{calculating loss expectations}
\For{$\mathrm{q}=1\;\mathrm{\mathbf{to}}\;Q$}{
sample $\mathbf{r}$ from $\mathcal{N}(\mathbf{0},\mathbf{1})$\; 
sample $\boldsymbol{\Gamma}$ from a multinomial distribution with probability $\{\omega_k\}_{k=1}^K$ \;
calculate $\mathcal{L}_\text{adv}$ and $\mathcal{L}_\text{entropy}$ by Eq.~\eqref{eq:L_adv_1}, Eq.~\eqref{eq:L_adv_2} and Eq.~\eqref{eq:L_entropy_2}\;
$\mathcal{L} \leftarrow \mathcal{L}+\frac{1}{Q}(\mathcal{L}_\text{adv}-\lambda\cdot\mathcal{L}_\text{entropy})$\;
}
\tcc{updating network weights}
update $\mathbf{W}$ with stochastic gradient ascent $\mathbf{W}\leftarrow \mathbf{W}+\nabla_{\mathbf{W}}\mathcal{L}$\;
}}
\KwOut{optimized network $F_{\mathbf{W}^*}$}
\end{algorithm}

\subsection{Black-Box Attacks with Distributional Prior}
\label{sec:Black-Box Attack with Distributional Prior}
To leverage the transferability of adversarial distribution priors, we further design two query-based attack methods based on the transfer-prior against black-box models, which involve fine-tuning for individual attack locations (Sec.~\ref{sec:LO}) and fine-tuning the parameters of the adversarial distribution  (Sec.~\ref{sec:DT}). Details are shown in the following.

\subsubsection{Location-Optimization Attack (DOP-LOA)} \label{sec:LO}
To search for an effective adversarial location with limited information, the general idea is to directly sample alternative locations from the distribution prior or use a grid traversal to search for feasible attack locations within the distribution prior range. However, these methods lack heuristics and result in low search efficiency. Additionally, they cannot overcome variations in the adversarial distribution between surrogate and black-box models. To address these issues, we propose to use the Bayesian Optimization Algorithm (BOA)~\cite{snoek2012practical} to better break the dilemma between exploration and exploitation for searching worst-case placement locations. BOA is an effective method to solve global optimization problems. It comprises two key components: a surrogate model (e.g., Gaussian Process) is employed to model the unknown objective function, and the acquisition function is utilized to determine the next sampling point from the Bayesian posterior. The selection of the acquisition function is critical as it balances the exploration and exploitation dilemmas in the optimization process.

To rationally apply the distribution prior, we limit the range of Bayesian optimization to the $3\sigma$ range around the distribution prior center $\mu$. Specifically, we use a Gaussian Process (GP) as the surrogate model to fit the Bayesian posterior distribution of the unknown function $\mathcal{L}_\mathrm{adv}(\boldsymbol{\theta})$ under observation samples $\mathcal{O}_N=\{\boldsymbol{\theta}_i, \mathcal{L}_{\mathrm{adv}}(\boldsymbol{\theta}_i)\}_{i=1}^N$: 

\begin{equation}
p(\boldsymbol{\theta}_{s+1}|\mathcal{O}_{N+s}) \sim \mathcal{N}(\tilde{\mu}_s(\boldsymbol{\theta}_{s+1}), \tilde\sigma^2_s(\boldsymbol{\theta}_{s+1}) ),
\end{equation}
where $s$ represents the number of iterations, $\tilde{\mu}_s$ and $\tilde{\sigma}^2_s$ is the mean and variance of the marginal distribution. Now, we can estimate the distribution that $\boldsymbol{\theta}_{s+1}$ obeyed at any value. Then, we use the Upper Confidence Bound of GP (GP-UCB) as the acquisition function (AC), which takes a weighted average of the mean and variance, and obtain the next sampling point by maximizing the AC:
\begin{equation}
\begin{split}
\boldsymbol{\theta}_{s+1} &\leftarrow \arg \max_{\boldsymbol{\theta}_{s+1}} \alpha_{\mathrm{GP-UCB}}(\boldsymbol{\theta}_{s+1}|\mathcal{O}_{N+s}), \\
\mathrm{where} \;
\alpha_{\mathrm{GP-UCB}}&(\boldsymbol{\theta}_{s+1}|\mathcal{O}_{N+s}) \leftarrow \tilde{\mu}_s(\boldsymbol{\theta}_{s+1}) + \beta \cdot \tilde{\sigma_s}(\boldsymbol{\theta}_{s+1}),
\end{split}
\end{equation}
where $\beta$ is a hyperparameter for balance exploration and exploitation. The overall algorithm is shown in Appendix, based on the objective function, we first obtain a batch of observation samples within the limited search space (i.e., $3\sigma$ space of the adversarial distribution prior) and initialize the GP model. In each iteration, we first maximize the AC to obtain the best next sampling point, add it to the observation sample set, and then update the GP model. Through multiple iterations of BOA, we heuristically explore the best attack location within the distribution prior. 

% \begin{equation}
% \mathcal{L}(\boldsymbol{\theta})=\left\{
% 	\begin{aligned}
% &\mathcal{L}_\text{CE}\bigl ( f \bigl (\mathcal{G}(\mathbf{X},\mathbf{P}, \boldsymbol{\theta} \bigr ),t \bigr ), &\;\mathrm{in\; classification};\\
% &1-\mathrm{cos}\bigl ( f \bigl (\mathcal{G}(\mathbf{X},\mathbf{P}, \boldsymbol{\theta} \bigr ),f(\mathbf{X}) \bigr ), &\;\mathrm{in\; identification}.\\
% 	\end{aligned}
% 	\right
% 	.
% \end{equation}

\subsubsection{Distribution-Transfer Attack (DOP-DTA)} \label{sec:DT}

Despite the ability of DOP-LOA to obtain specific adversarial locations under black-box models, it only generates a single adversarial sample per optimization process and cannot generalize to adversarial regions, which contradicts our original intention of using distribution modeling. To address this issue, we propose DOP-DTA, which aims to transfer the distribution prior from the surrogate model to the black-box model, thus enabling us to explore the model's vulnerabilities despite having limited information while enhancing the performance of the attack. Therefore, we further optimize the distribution parameters under the black-box settings. Formally, DOP-DTA also aims to address Eq.\eqref{eq:max problem_2} by utilizing the distribution prior generated by $F_{\mathbf{W}^*}$. However, the model's structure and gradient information are unavailable now. Thus, we employ Natural Evolution Strategies (NES) to obtain the natural gradient of the adversarial loss w.r.t the distribution parameters. For the entropy regularization loss, we directly provide its analytical expression to compute the gradient. Accordingly, we can derive the total gradients of the objective function Eq.\eqref{eq:max problem_2} w.r.t the distribution parameters, i.e., the $\omega_k$, $\boldsymbol{\mu}_k$, and $\boldsymbol{\sigma}_k$, as follows:
\begin{equation}\small
    \begin{split}
    \nabla_{\omega_k} = \mathbb{E}_{\mathcal{N}(\mathbf{r}|\mathbf{0},\mathbf{I})} & \left \{   \gamma_k \cdot \left [\mathcal{L}_\text{adv} \cdot \frac{1}{\omega _k} - \lambda \right ]\right \}, \\
    \nabla_{\boldsymbol{\mu}_k} = \mathbb{E}_{\mathcal{N}(\mathbf{r}|\mathbf{0},\mathbf{I})}& \left \{\gamma _k \cdot \left [\mathcal{L}_\text{adv} \cdot \frac{\boldsymbol{\sigma}_k \mathbf{r} }{\omega _k} \!- \!\lambda \!\cdot\! 2 \tanh(\frac{\boldsymbol{\mu} _k+\boldsymbol{\sigma}_k \mathbf{r}}{\tau})\cdot\frac{1}{\tau}\right]\!\right \}\!, \\
    \nabla_{\boldsymbol{\sigma}_k} = \mathbb{E}_{\mathcal{N}(\mathbf{r}|\mathbf{0},\mathbf{I})}&\left \{\gamma _k \cdot \left [\mathcal{L}_\text{adv} \cdot \frac{\boldsymbol{\sigma_k} (\mathbf{r}^2-1) }{2\omega_k} \right.\right.\\
    &\left.\left. + \lambda \cdot \frac{(1-2\cdot \tanh(\frac{\boldsymbol{\mu} _k+\boldsymbol{\sigma}_k \mathbf{r}}{\tau}  )\cdot \frac{\boldsymbol{\sigma}_k \mathbf{r}}{\tau})}{\boldsymbol{\sigma}_k} \right] \right \}.
    \end{split}
    \label{eq:DOP-DT}
\end{equation}
The proof of Eq.~\eqref{eq:DOP-DT} and the algorithm of DOP-DTA will be detailed in the Appendix. We use iterative gradient ascent to optimize the distribution parameters of each component. For each image, we first obtain the parameters of the multi-modal distribution from the distribution mapping network as the initial values for optimization. Then, we iteratively perform gradient ascent to optimize the distribution parameters of each Gaussian component. As it is necessary to ensure that $ {\textstyle \sum_{k=1}^{K}}\omega_k = 1 $, we perform normalization on $\omega_k$ after each iteration.

\subsection{Distribution-Based Adversarial Training}
\label{sec:Distribution-Based Adversarial Training}
The concept of adversarial training (AT) traces back to the pioneering work of Goodfellow et al.\cite{goodfellow2014explaining}, which is widely regarded as the most efficient approach to improving the robustness of deep learning models. Although prior work attempts to employ AT to bolster the models' robustness against location-aware adversarial patches\cite{wei2022adversarial,guo2021meaningful}, they incur an unacceptable overhead in producing a diverse set of adversarial samples. We posit that DOPatch can alleviate this issue owing to its ability to capture the distribution of adversarial locations, which endows it with the capacity to generate effective attacks and makes it a natural fit for AT. In this section, we discuss using DOPatch for distribution-based adversarial training (DOP-DMAT).

\begin{algorithm}
\footnotesize
\caption{\footnotesize Algorithms for DOP-DMAT}\label{al:dopat}
\KwIn{dataset $\mathcal{D}$, distribution mapping network $F_\mathbf{W}$, target model $f_{\boldsymbol{\hbar}}$, training epochs $N$, components number $K$, Monte Carlo sample number $Q$, balance hyperparameter $\lambda$.}
Initialize $\mathbf{W}$ and $\boldsymbol{\hbar}$\;
$\{\omega_k\}_{k=1}^K \leftarrow \frac{1}{K}$\;
\For{$\mathrm{epoch}=1\;\mathbf{to}\; N$}{
\For{$\mathrm{each\;minbatch} \;\mathcal{B}\subset\mathcal{D}$}{
\tcc{inner maximization:}
$\{\boldsymbol{\mu}_k^*, \boldsymbol{\sigma}_k^*\}_{k=1}^K \leftarrow F_\mathbf{W}(\mathcal{B})$ \;
$\mathcal{L} \leftarrow 0$\;
\For{$\mathrm{q}=1\;\mathrm{\mathbf{to}}\;Q$}{
$\mathcal{L} \leftarrow \mathcal{L}+\frac{1}{Q}(\mathcal{L}_\text{adv}(\mathcal{B})-\lambda\cdot\mathcal{L}_\text{entropy})$\;
}
$\mathbf{W}\leftarrow \mathbf{W}+\nabla_{\mathbf{W}}\mathcal{L}$\;
\tcc{outer minimization:}
sample adversarial location $\boldsymbol{\theta}$ according to Eq.~\eqref{eq:parameterized} \;
generate a set of  $(\textbf{X}^{\mathrm{adv}},y^{\prime})\in \mathcal{B}^{\mathrm{adv}}$\;
$\boldsymbol{\hbar}\leftarrow \boldsymbol{\hbar}+\nabla_{\boldsymbol{\hbar}}\mathcal{L_\mathrm{adv}}([\mathcal{B}^{\mathrm{adv}}, \mathcal{B}])$\;
}}
\KwOut{improved target model $f_{\boldsymbol{\hbar}}$}
\end{algorithm}

Formally, DOP-DMAT is formulated as a distribution-based min-max optimization problem: 
\begin{equation}
\min_{\boldsymbol{\hbar}}\max_{\mathbf{W}} \left [ \mathbb{E}_{\mathcal{N}(\mathbf{r}|\mathbf{0},\mathbf{I}))}\left (\mathcal{L}_\text{adv} - \lambda\cdot \mathcal{L}_\text{entropy} \right )  \right ],
\label{eq:min-max problem}
\end{equation}
where $\boldsymbol{\hbar}$ is the weight of the target model $f_{\boldsymbol{\hbar}}(\cdot)$. As shown in Eq.~\eqref{eq:min-max problem}, we use DOPatch to solve the inner maximization problem, seeking the worst-case distribution of adversarial patch locations while maximizing both the adversarial loss and entropy regularization loss. The outer minimization aims to optimize the weights of the target model to minimize the expected loss on the worst-case distribution. Solving min-max problems directly is usually challenging. Based on Danskin's theorem~\cite{danskin2012theory}, a general solution is first to solve the inner problem to generate adversarial samples and then use gradient descent to solve the outer problem. For DOP-DMAT, we first optimize the distribution mapping network $F_{\mathbf{W}}$ in each iteration, then sample alternative locations using Monte Carlo sampling after obtaining the current adversarial distribution. We then feed the generated adversarial patch samples with clean samples to update the weights of target model $f_{\boldsymbol{\hbar}}$. Algorithm~\ref{al:dopat} shows the training process of DOP-DMAT in a single iteration cycle. 
% Additionally, as DOP-DMAT belongs to the distribution-based adversarial training methods, its convergence is theoretically proven by Dong et al.~\cite{dong2020adversarial}. We will also verify the convergence performance in experiments.

\section{Experiments}
\subsection{Experimental Settings} \label{sec:setting}
\noindent \textbf{Datasets.} We evaluate our method on various public face datasets: LFW~\cite{huang2008labeled} and CelebA~\cite{liu2018large}. These datasets contain face images from 5749 and 8192 identities, respectively, and all images are used to construct the corresponding identity databases. We randomly select 1000 images for each dataset to perform impersonation attacks.  

\noindent \textbf{Target Models.} We conduct experiments on various face recognition models, which span different backbone structures and training paradigms: FaceNet~\cite{schroff2015facenet} with Inception-ResNet-v2, ArcFace~\cite{deng2019arcface} with IResNet-18, CosFace~\cite{wang2018cosface} with IResNet-18, SphereFace\cite{liu2017sphereface} with IResNet-50 and MobileFaceNet~\cite{chen2018mobilefacenets} with MobileNet-v2. We obtain the pre-training model separately from open-source projects\footnote[1]{\scriptsize \href{https://github.com/timesler/facenet-pytorch}{\textcolor{black}{https://github.com/timesler/facenet-pytorch}}}~\footnote[2]{\scriptsize\href{https://github.com/deepinsight/insightface/tree/master/recognition}{\textcolor{black}{https://github.com/deepinsight/insightface/tree/master/recognition}}}~\footnote[3]{\scriptsize\href{https://github.com/wujiyang/Face_Pytorch}{\textcolor{black}{https://github.com/wujiyang/Face\_Pytorch}}}~\footnote[4]{\scriptsize\href{https://github.com/Xiaoccer/MobileFaceNet_Pytorch}{\textcolor{black}{https://github.com/Xiaoccer/MobileFaceNet\_Pytorch}}}, each
model obtain over 99\% verification accuracy on LFW image pairs by following its optimal threshold. Following the settings in \cite{wei2022adversarial} and \cite{guo2021meaningful}, we utilize these models to perform the attack on face classification and identification tasks. For the identification task, we use the pre-trained model as a feature extractor to obtain embeddings of face images, then select its nearest neighbor among all identities in the dataset as the prediction result. For the classification task, we add a fully connected layer after the aforementioned open-source models and fine-tune it on the target dataset until convergence. Then, we use the trained classifier to obtain the top-1 prediction for face images.

\noindent \textbf{Compared methods.} As our method improves location-aware patch attacks, we select the current state-of-the-art method, RHDE~\cite{wei2022adversarial,guo2021meaningful}, as the primary baseline. We also choose GDPA~\cite{li2021generative} as another comparative baseline and set a fixed pattern for GDPA while only optimizing for adversarial locations. In addition, two potential baselines are also considered, where adversarial locations are the mean location on the distribution optimized by DOPatch and directly transferred to attack other black-box models. This operation actually perform a transfer-based attack, which is named \textbf{DOP}. Based on DOPatch, we can also conduct the query-based attack, but a random sample from the distribution prior, which is named \textbf{DOP-Rd}. For adversarial training, we mainly compare standard training, data augmentation using randomly placed patches (Random-Aug), and a SOTA method AT method for adversarial patches \cite{li2021generative}.

\noindent \textbf{Evaluation metrics.} Regarding the attacks, we use the attack success rate (\textbf{ASR}) and the number of queries (\textbf{NQ}) as evaluation metrics. The former refers to the percentage of successfully attacked samples within the query limit among all the attacked samples. The latter refers to the average number of queries for all attacked samples. For adversarial training, we evaluate the attack success rate of the model under five location-aware patch methods.

\noindent \textbf{Implementation Details.} In experiments, we set the default values for the Gaussian component number as $K=5$ and the entropy regularization strength as $\lambda=0.1$. The effects of these parameters are reported in Sec.~\ref{sec:ablation}. We employ FaceNet, SphereFace, and MobileFaceNet as the surrogate models used in the first-stage optimization, respectively, for obtaining the adversarial distribution prior. The remaining models served as the target models for the black-box attacks. Specifically, for FaceNet and MobileNet, we utilize a combination structure of three residual blocks following the approach proposed in\cite{zhu2017unpaired} as the feature extraction backbone for the distribution mapping network. While for SphereFace, we utilize the ResNet-50~\cite{he2016deep} pre-trained on the CASIA-Webface dataset as the backbone. During the distribution mapping network training, we adopt the Adam optimizer with an initial learning rate of 0.01 for training 500 epochs, and the learning rate decayed by a factor of 0.2 for every 150 epochs. In the DOP-LOA algorithm, the initial number of samples for Bayesian optimization is set to 10, and the maximum number of queries is set to 500. For the DOP-DTA algorithm, we set the learning rate of NES to 100, the population size to 10, and the maximum number of iterations to 50. Finally, the slope parameter of the $tanh(\cdot)$ function is set to $\tau=3000$.

% \textbf{(3)} The choice of surrogate models has a significant impact on the performance of DOPatch. In our experiments, we observed that using FaceNet as the surrogate model achieves the best attack performance on the remaining black-box models. We attribute this to the superior cross-model transfer capability of the adversarial distribution associated with FaceNet, allowing it to generalize across multiple models' adversarial regions.

\subsection{Ablation Studies} \label{sec:ablation}

\begin{table*}[t]
\caption{The attack results of our method under different fixed patterns are presented. We select various cartoon stickers, the pattern generated by LaVAN~\cite{karmon2018lavan} and random Gaussian noise as fixed patterns. By using FaceNet as the surrogate model, we perform black-box attacks on the remaining models. The attack success rate (ASR) and the number of queries (NQ) are reported.}
\setlength\tabcolsep{8pt}
\renewcommand\arraystretch{1.1}
\centering
\begin{tabular}{c|c|cc|cc|cc|cc|cc}
\hline
\multirow{2}{*}{Pattern} & \multirow{2}{*}{Method} & \multicolumn{2}{c|}{FaceNet} & \multicolumn{2}{c|}{SphereFace} & \multicolumn{2}{c|}{MobileFaceNet} & \multicolumn{2}{c|}{ArcFace}  & \multicolumn{2}{c}{CosFace}   \\ \cline{3-12} 
                         &                         & ASR~$\uparrow$           & NQ~$\downarrow$           & ASR~$\uparrow$            & NQ~$\downarrow$             & ASR~$\uparrow$              & NQ~$\downarrow$              & ASR~$\uparrow$           & NQ~$\downarrow$            & ASR~$\uparrow$           & NQ~$\downarrow$            \\ \hline
\multirow{3}{*}{\begin{minipage}[b]{0.10\columnwidth}
		\centering
		\raisebox{-.5\height}{\includegraphics[width=\linewidth]{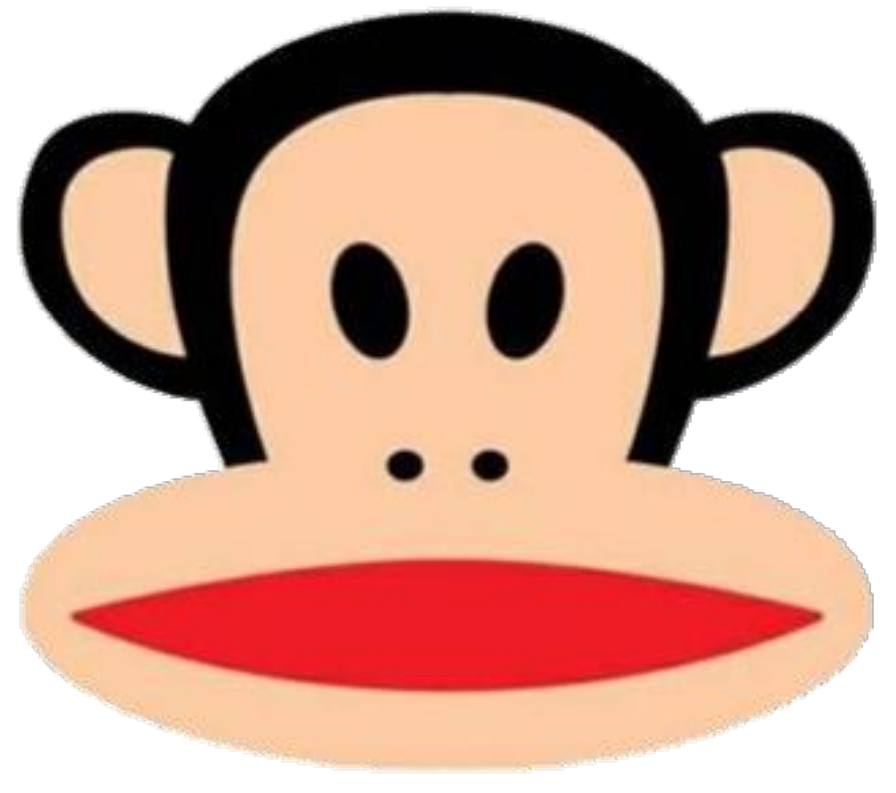}}
	\end{minipage}}        & DOP-Rd                  & 83.5\%          & 219          & 36.6\%           & 1783           & 83.4\%             & 239             & 28.1\%          & 2617          & 35.7\%          & 1863          \\
                         & DOP-LOA                 & \textbf{84.3\%} & \textbf{110} & 44.5\%           & 640            & 90.6\%             & 73              & 36.7\%          & 895           & 47.4\%          & 579           \\
                         & DOP-DTA                 & 83.0\%          & 142          & \textbf{61.9\%}  & \textbf{392}   & \textbf{96.2\%}    & \textbf{71}     & \textbf{53.2\%} & \textbf{571}  & \textbf{65.5\%} & \textbf{371}  \\ \hline
\multirow{3}{*}{\begin{minipage}[b]{0.12\columnwidth}
		\centering
		\raisebox{-.5\height}{\includegraphics[width=\linewidth]{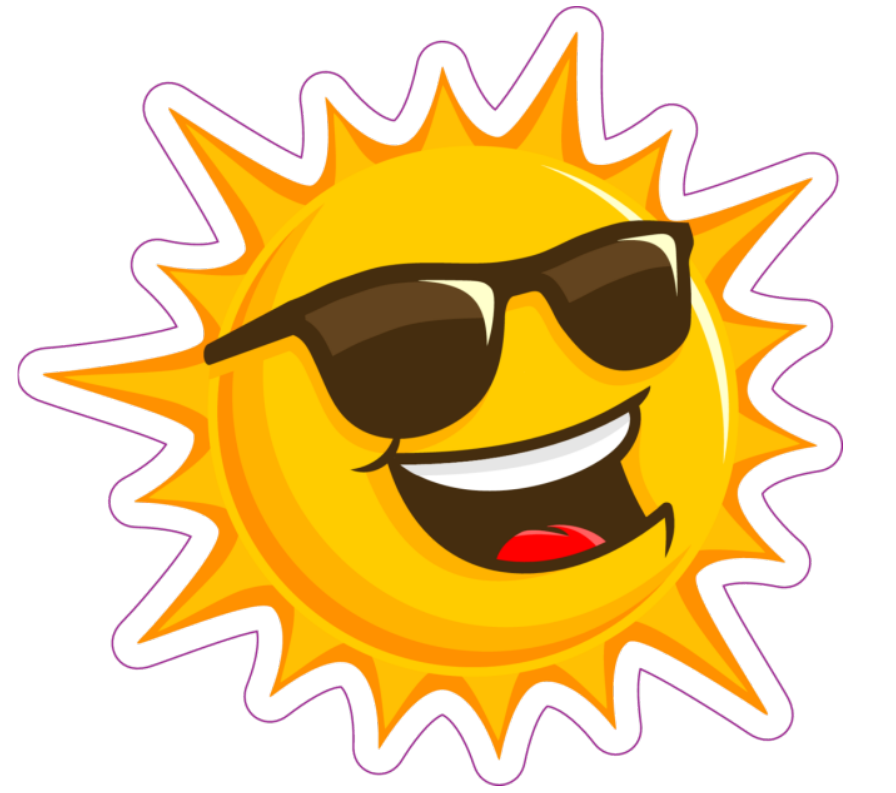}}
	\end{minipage}}        & DOP-Rd                  & \textbf{88.2\%} & 167          & 55.5\%           & 870            & 95.1\%             & 90              & 46.1\%          & 1274          & 56.3\%          & 867           \\
                         & DOP-LOA                 & 78.9\%          & \textbf{149} & 46.1\%           & 609            & 96.9\%             & \textbf{31}     & 39.1\%          & 817           & 43.0\%          & 710           \\
                         & DOP-DTA                 & 79.9\%          & 167          & \textbf{62.2\%}  & \textbf{388}   & \textbf{97.1\%}    & 70              & \textbf{52.7\%} & \textbf{563}  & \textbf{62.6\%} & \textbf{377}  \\ \hline
\multirow{3}{*}{\begin{minipage}[b]{0.11\columnwidth}
		\centering
		\raisebox{-.5\height}{\includegraphics[width=\linewidth]{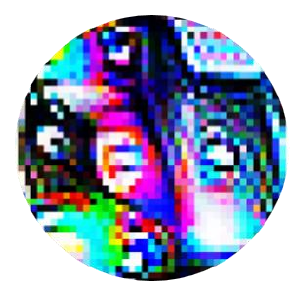}}
	\end{minipage}}        & DOP-Rd                  & \textbf{82.9\%} & 226          & 27.1\%           & 2764           & 38.8\%             & 1670            & 43.6\%          & 1338          & 91.2\%          & 128           \\
                         & DOP-LOA                 & 82.8\%          & \textbf{115} & 30.5\%           & 1171           & 39.8\%             & 814             & 46.1\%          & 597           & 94.5\%          & \textbf{42}   \\
                         & DOP-DTA                 & 79.3\%          & 169          & \textbf{49.1\%}  & \textbf{648}   & \textbf{64.8\%}    & \textbf{359}    & \textbf{66.6\%} & \textbf{320}  & \textbf{96.7\%} & 63            \\ \hline
\multirow{3}{*}{\begin{minipage}[b]{0.11\columnwidth}
		\centering
		\raisebox{-.5\height}{\includegraphics[width=\linewidth]{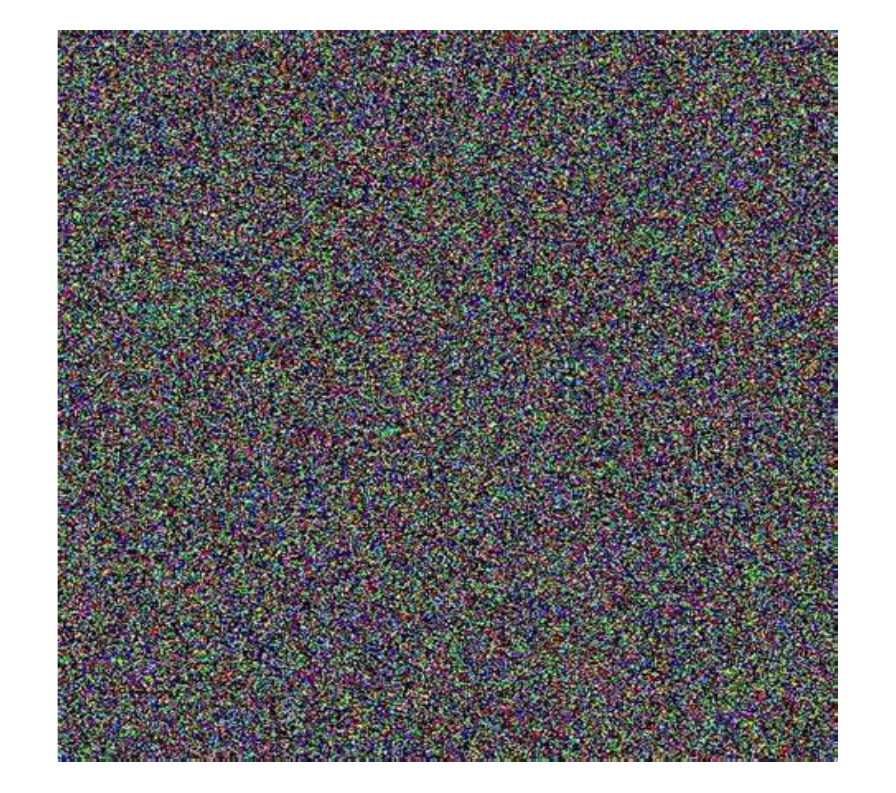}}
	\end{minipage}}        & DOP-Rd                  & 35.6\%          & 1812         & 18.6\%           & 4377           & 58.3\%             & 721             & 25.5\%          & 2933          & 20.8\%          & 3811          \\
                         & DOP-LOA                 & 30.5\%         & 1141         & 25.0\%           & 1505           & 55.5\%             & 407             & 31.3\%          & \textbf{1104} & 23.4\%          & 1640          \\
                         & DOP-DTA                 & \textbf{72.2\%} & \textbf{392} & \textbf{37.9\%}  & \textbf{1094}  & \textbf{86.1\%}    & \textbf{210}    & \textbf{31.7\%} & 1204          & \textbf{29.9\%} & \textbf{1405} \\ \hline
\end{tabular}
\label{tab:ablation patterns}
\end{table*}

\begin{figure}[htb]
  \centering
  \includegraphics[width=0.99\linewidth]{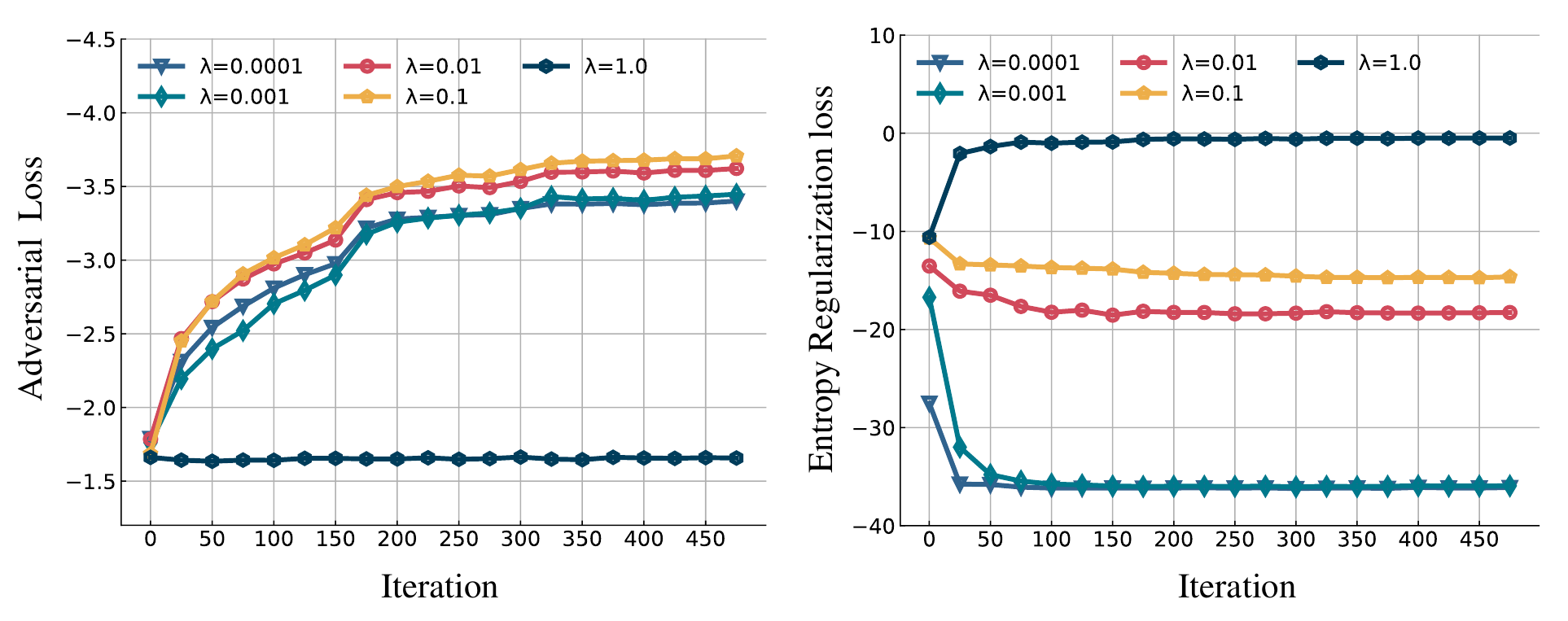}
   \caption{Convergence curves of the adversarial loss and the entropy regularization loss with different values of $\lambda$ ($1.0$, $0.1$, $0.01$, $0.001$, and $0.0001$) for FaceNet as the surrogate model.}
   \label{fig: ablation lambda} 
\end{figure}

% \subsubsection{Convergence under loss balance parameter}
\noindent \textbf{Convergence under different $\lambda$.} For the optimization process of the distribution prior, i.e., in  DOPatch, we conduct experiments to investigate the convergence of the distribution mapping network under different loss balance parameters $\lambda$. We plot the adversarial and entropy regularization loss curves during iteration, as shown in Fig.~\ref{fig: ablation lambda}. Using FaceNet as the surrogate model, the distribution mapping network converges effectively when $\lambda<1.0$, and the adversarial loss converges to the maximum when $\lambda=0.1$ and $0.01$, indicating the optimized distribution prior is best in attack performance. Furthermore, a larger entropy regularization loss indicates that DOPatch can capture a more comprehensive adversarial region, which improves the performance of transfer attacks. In conclusion, we consider DOPatch to exhibit the best convergence performance when lambda is set to 0.1.

% \subsubsection{Effect of Different Surrogate Models}

\begin{figure}[t]
  \centering
  \includegraphics[width=0.99\linewidth]{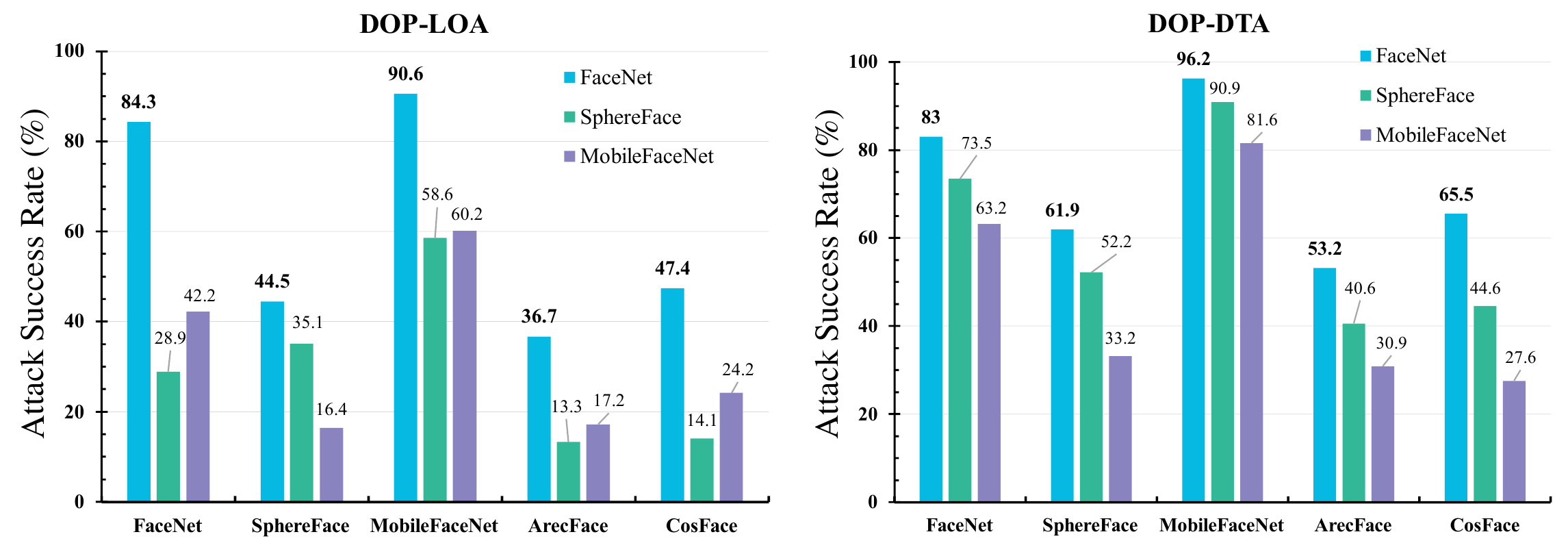}
   \caption{\textbf{Attack success rates} of DOP-LOA and DOP-DTA against different face classifiers. FaceNet, SphereFace, and MobileFaceNet serve as surrogate models, respectively.}
   \label{fig: ablation surrogate} 
\end{figure}

\noindent \textbf{Effect of different surrogate models.}
We utilize FaceNet, SphereFace, and MobileFaceNet as surrogate models, respectively, to investigate the impact of the obtained distribution prior on the attack performance across different surrogate models. Fig.~\ref{fig: ablation surrogate} presents the attack success rates of DOP-LOA and DOP-DTA with adversarial distribution priors using different surrogate models. The results demonstrate that using FaceNet as the surrogate model achieves the best performance, which can be attributed to the superior cross-model transfer capability of the adversarial distribution associated with FaceNet, allowing it to generalize across multiple models' adversarial regions.

\begin{figure*}[ht]
  \centering
  \includegraphics[width=0.95\linewidth]{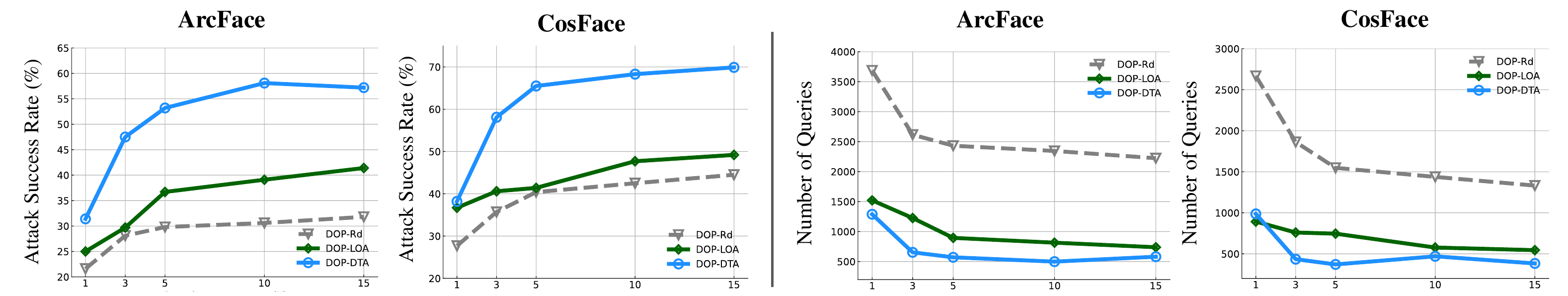}
   \caption{\textbf{Attack success rate} and the \textbf{number of query} curves for DOP-Rd, DOP-LOA, and DOP-DTA using $K$ ranging from $0$ to $15$.}
   \label{fig: ablation K} 
\end{figure*}

% \subsubsection{Effect of Gaussian Component Number}
\noindent \textbf{Effect of different $K$ values.}
DOPatch parametrizes the adversarial region as a multimodal Gaussian distribution to capture a more comprehensive set of adversarial locations. Therefore, we further investigate the impact of different Gaussian component numbers on the performance of the proposed method. We set the number of components to $K\!\!=\!\!1, 3, 5, 10$, and $15$, then optimize the distribution prior under the LFW dataset and use FaceNet as the surrogate model. Fig.~\ref{fig: ablation K} demonstrates the attack success rates and query numbers of DOP-Rd, DOP-LOA, and DOP-DTA against ArcFace and CosFace in the face classification task. We observe that for the proposed query-based attack methods, the ASR increases while the corresponding NQ decreases as the number of Gaussian components $K$ used in optimizing the distribution prior increases. This phenomenon is particularly prominent for $K=1$ to $5$ and becomes less pronounced for $K>5$, where ASR and NQ show relatively smooth variations w.r.t $K$ increases. Furthermore, among different $K$ settings, the proposed DOP-DTA achieves the best performance in terms of ASR and NQ, followed by DOP-LOA, while both methods outperform DOP-Rd significantly.

\begin{table*}[t]
\caption{Attack performance evaluation on the LFW and CelebA datasets against \textbf{face classification} models. The table presents the attack success rate (ASR) and number of queries (NQ) using different location-aware methods against FaceNet, SphereFace, MobileFaceNet, ArcFace, and CosFace.}
\setlength\tabcolsep{1.4pt}
\renewcommand\arraystretch{1.4}
\centering
\begin{tabular}{c|cccccccccc|cccccccccc}
\hline
\multirow{3}{*}{Method} & \multicolumn{10}{c|}{LFW-1000subset~\cite{huang2008labeled}}                                                                                                                                       & \multicolumn{10}{c}{CelebA-1000subset~\cite{liu2018large}}                                                                                                                                    \\ \cline{2-21} 
                        & \multicolumn{2}{c|}{\scriptsize{FaceNet}}           & \multicolumn{2}{c|}{\scriptsize{SphereFace}}     & \multicolumn{2}{c|}{\scriptsize{MobileFaceNet}}        & \multicolumn{2}{c|}{\scriptsize{ArcFace}}           & \multicolumn{2}{c|}{\scriptsize{CosFace}} & \multicolumn{2}{c|}{\scriptsize{FaceNet}} & \multicolumn{2}{c|}{\scriptsize{SphereFace}} & \multicolumn{2}{c|}{\scriptsize{MobileFaceNet}} & \multicolumn{2}{c|}{\scriptsize{ArcFace}} & \multicolumn{2}{c}{\scriptsize{CosFace}} \\ \cline{2-21} 
                        & \scriptsize{ASR~$\uparrow$}  & \multicolumn{1}{c|}{\scriptsize{NQ~$\downarrow$}}   & \scriptsize{ASR~$\uparrow$}  & \multicolumn{1}{c|}{\scriptsize{NQ~$\downarrow$}}   & \scriptsize{ASR~$\uparrow$}    & \multicolumn{1}{c|}{\scriptsize{NQ~$\downarrow$}}   & ASR  & \multicolumn{1}{c|}{\scriptsize{NQ~$\downarrow$}}   & \scriptsize{ASR~$\uparrow$}           & \scriptsize{NQ~$\downarrow$}           & \scriptsize{ASR~$\uparrow$}  & \multicolumn{1}{c|}{\scriptsize{NQ~$\downarrow$}}  & \scriptsize{ASR~$\uparrow$}  & \multicolumn{1}{c|}{\scriptsize{NQ~$\downarrow$}}  & \scriptsize{ASR~$\uparrow$}    & \multicolumn{1}{c|}{\scriptsize{NQ~$\downarrow$}}   & \scriptsize{ASR~$\uparrow$}  & \multicolumn{1}{c|}{\scriptsize{NQ~$\downarrow$}}  & \scriptsize{ASR~$\uparrow$}           & \scriptsize{NQ~$\downarrow$}          \\ \hline
Clean                   & 0.1\%  & \multicolumn{1}{c|}{-}     & 0.0\%  & \multicolumn{1}{c|}{-}     & 0.2\%    & \multicolumn{1}{c|}{-}     & 1.1\%  & \multicolumn{1}{c|}{-}     & 0.8\%           &-              & 6.6\%     & \multicolumn{1}{c|}{-}    & 13.3\%      & \multicolumn{1}{c|}{-}    & 13.3\%       & \multicolumn{1}{c|}{-}     &  13.9\%    & \multicolumn{1}{c|}{-}    &  8.3\%             &-             \\ \hline
GDPA~\cite{li2021generative}                    & 53.0\% & \multicolumn{1}{c|}{-}     & 30.4\% & \multicolumn{1}{c|}{-}     & 57.7\%   & \multicolumn{1}{c|}{-}     & 25.5\% & \multicolumn{1}{c|}{-}     & 20.6\%          &-              & 49.8\%     & \multicolumn{1}{c|}{-}    & 47.6\%     & \multicolumn{1}{c|}{-}    & 57.3\%       & \multicolumn{1}{c|}{-}     & 42.8\%     & \multicolumn{1}{c|}{-}    & 38.6\%              &-             \\
RHDE~\cite{wei2022adversarial}                    & 42.2\% & \multicolumn{1}{c|}{1945} & 31.3\% & \multicolumn{1}{c|}{2950} & 65.9\%   & \multicolumn{1}{c|}{801}  & 43.7\% & \multicolumn{1}{c|}{1743} & 61.6\%          & 920          & 74.3\%     & \multicolumn{1}{c|}{631}    &  74.6\%     & \multicolumn{1}{c|}{558}    & 72.9\%       & \multicolumn{1}{c|}{621}     & 61.4\%     & \multicolumn{1}{c|}{894}    & 65.5\%              & 796            \\ \hline
DOP                  & 72.6\% & \multicolumn{1}{c|}{-}  & 23.1\% & \multicolumn{1}{c|}{-} & 52.4\%   & \multicolumn{1}{c|}{-}  & 12.7\% & \multicolumn{1}{c|}{-} & 17.5\%          & -         & 50.2\% & \multicolumn{1}{c|}{-} & 40.2\% & \multicolumn{1}{c|}{-} & 56.9\%   & \multicolumn{1}{c|}{-}  & 40.4\% & \multicolumn{1}{c|}{-} & 37.5\%          & -         \\

DOP-Rd                  & 83.5\% & \multicolumn{1}{c|}{219}  & 36.6\% & \multicolumn{1}{c|}{1783} & 83.4\%   & \multicolumn{1}{c|}{239}  & 28.1\% & \multicolumn{1}{c|}{2617} & 35.7\%          & 1863         & 89.3\% & \multicolumn{1}{c|}{149} & 77.6\% & \multicolumn{1}{c|}{327} & 91.5\%   & \multicolumn{1}{c|}{118}  & 74.5\% & \multicolumn{1}{c|}{388} & 64.0\%          & 601         \\
DOP-LOA                 & \textbf{84.3\%} & \multicolumn{1}{c|}{\textbf{110}}  & 44.5\% & \multicolumn{1}{c|}{639}  & 90.6\%   & \multicolumn{1}{c|}{73}   & 36.7\% & \multicolumn{1}{c|}{895}  & 47.4\%          & 579          & \textbf{93.0\%} & \multicolumn{1}{c|}{\textbf{61}}  & 78.9\% & \multicolumn{1}{c|}{\textbf{145}} & \textbf{92.2\%}   & \multicolumn{1}{c|}{\textbf{55}}   & 77.3\% & \multicolumn{1}{c|}{\textbf{168}} & 63.3\%          & 312         \\
DOP-DTA                 & 83.0\% & \multicolumn{1}{c|}{142}  & \textbf{61.9\%} & \multicolumn{1}{c|}{\textbf{392}}  & \textbf{96.2\%}   & \multicolumn{1}{c|}{\textbf{71}}   & \textbf{53.2\%} & \multicolumn{1}{c|}{\textbf{571}}  & \textbf{65.5\%}          & \textbf{371}          & 82.2\% & \multicolumn{1}{c|}{145} & \textbf{79.6\%} & \multicolumn{1}{c|}{178} & 90.2\%   & \multicolumn{1}{c|}{91}   & \textbf{78.7\%} & \multicolumn{1}{c|}{196} & \textbf{69.5\%}          & \textbf{286}         \\ \hline
\end{tabular}
\label{tab:classification}
\end{table*}

\begin{figure*}[thb]
  \centering
  \includegraphics[width=0.99\linewidth]{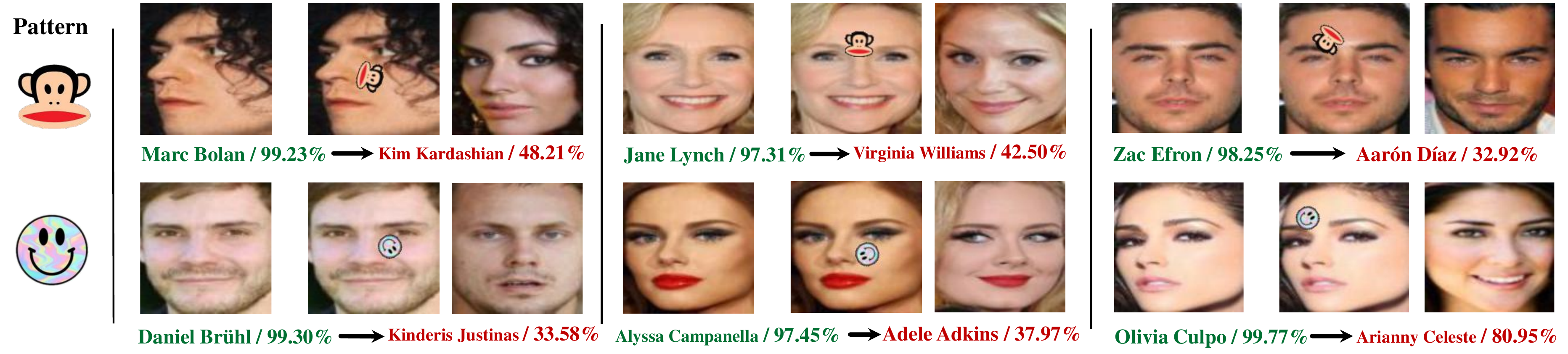}
   \caption{Examples of attacks using various patterns. Each group showcases the original face image, the face image post-attack, and the face image of the misclassified class. The correct identity is highlighted in green text, while the predicted wrong identity after the attack is marked in red text. The accompanying percentages indicate the confidence values of the predicted results.}
   \label{fig:vis1} 
\end{figure*}

% \subsubsection{Performance of Different Patterns}
\noindent \textbf{Effect of different patterns.} We further conducted experiments using patches with different patterns. We selected three types of patterns: cartoon stickers, adversarial patches generated by LaVAN~\cite{karmon2018lavan}, and the Gaussian random noise. Table~\ref{tab:ablation patterns} present the ASR and NQ of the transfer attacks against various face classifiers using FaceNet as the surrogate model. The results demonstrate the effectiveness of DOPatch in conducting attacks with different patterns. Moreover, meaningful patterns (cartoon stickers) yield better attack performance than meaningless textures (adversarial patches and Gaussian noise). Additionally, in most cases, DOP-DTA exhibits the best attack performance.

\subsection{Attack Results on Face Classification Task}

Table~\ref{tab:classification} presents the experimental results on the face classification task, where we compare the state-of-the-art methods, GDPA and RHDE. For our proposed methods (i.e., DOP, DOP-Rd, DOP-LOA, and DOP-DTA), we utilize FaceNet as the surrogate model. Based on the results, We have the following findings: \textbf{1)} The proposed DOPatch method achieves superior attack performance compared to both the white-box attack method (GDPA) and the black-box attack method (RHDE). It achieves the highest attack success rate against all five face classification models on both LFW and CelebA datasets. Moreover, compared to RHDE, DOPatch significantly improves attack efficiency with fewer required attack queries, which is mainly attributed to the ability of DOPatch to learn the underlying distribution of adversarial locations from the surrogate model and leverage its significant transferability. \textbf{2)} Compared to random sampling from the distribution prior (DOP-Rd), the proposed transfer-attack methods, Location Optimization Attack (DOP-LOA), and Distribution Transfer Attack (DOP-DTA) achieve higher attack success rates. However, DOP-LOA notably improves attack efficiency by reducing the number of queries while maintaining a high attack performance. In contrast, DOP-DTA aims to transfer the distribution prior to black-box models, resulting in higher attack success rates but requiring more queries compared to DOP-LOA. We selected some visualization samples of attacks, which are shown in Fig.~\ref{fig:vis1}.

\subsection{Attack Results on Face Identification Task}
Face identification is a widely used technique in the face recognition task, which searches for the most similar face in the database to determine the identity of the input face image. To verify the proposed method, we conduct digital and physical attacks versus face identification, respectively. Specifically, we regard all the face images in LFW as the database and manipulate some query face images to perform impersonation attacks. The attack is successful if the returned identity is different from the query image. We adopt the same face identification setting with \cite{wei2022adversarial}. Table~\ref{tab:id attack} lists the experimental results of digital attacks against face identification using FaceNet, ArfFace, and CosFace. We can see that DOP-LOA and DOP-DTA significantly outperform RHDE for all the threat models versus ASR and NQ, showing that the distribution prior is indeed effective for query-based attacks. In addition, DOP-DTA is better than DOP-LOA, showing the superiority of distributional modeling.

\begin{table}[t]
\caption{Digital attack performance evaluation on the LFW dataset against \textbf{face identification} models. The table presents the attack success rate (ASR) and number of queries (NQ) using different location-aware methods against FaceNet, ArcFace and CosFace.}
\setlength\tabcolsep{5.3pt}
\renewcommand\arraystretch{1.1}
\centering

\begin{tabular}{c|cccccc}
\hline
\multirow{3}{*}{Method} & \multicolumn{6}{c}{LFW-1000subset~\cite{huang2008labeled}}                                                                                                                                                                  \\ \cline{2-7} 
                        & \multicolumn{2}{c|}{FaceNet}     & \multicolumn{2}{c|}{ArcFace}  & \multicolumn{2}{c}{CosFace} \\ \cline{2-7} 
                        & \scriptsize{ASR~$\uparrow$}  & \multicolumn{1}{c|}{\scriptsize{NQ~$\downarrow$}}   & \scriptsize{ASR~$\uparrow$}  & \multicolumn{1}{c|}{\scriptsize{NQ~$\downarrow$}}   & \scriptsize{ASR~$\uparrow$}  & \multicolumn{1}{c}{\scriptsize{NQ~$\downarrow$}}             \\ \hline
Clean                   & 0.1\%  & \multicolumn{1}{c|}{-}     & 0.1\% & \multicolumn{1}{c|}{-}     & 0.1\% & -                     \\ \hline
GDPA~\cite{li2021generative}                    & 10.9\% & \multicolumn{1}{c|}{-}     & 17.5\% & \multicolumn{1}{c|}{-}     &17.3\% & -                  \\
RHDE\cite{wei2022adversarial}                    & 48.8\%      & \multicolumn{1}{c|}{1666}     & 35.4\%      & \multicolumn{1}{c|}{2695}     & 42.9\%     &2136                    \\ \hline
DOP  & 31.5\% & \multicolumn{1}{c|}{-} & 13.1\% & \multicolumn{1}{c|}{-}  & 14.9\% &-                   \\
DOP-Rd                  & 56.1\% & \multicolumn{1}{c|}{839} & 38.0\% & \multicolumn{1}{c|}{1718} & 43.3\% &1409                   \\
DOP-LOA                 & 54.7\% & \multicolumn{1}{c|}{\textbf{447}} & 34.4\% & \multicolumn{1}{c|}{988} & 44.5\% &676                       \\
DOP-DTA                 & \textbf{56.1\%} & \multicolumn{1}{c|}{515} & \textbf{59.5\%} & \multicolumn{1}{c|}{\textbf{492}}  & \textbf{60.7\%} &\textbf{507}                        \\ \hline
\end{tabular}
\label{tab:id attack}
\end{table}

\begin{table*}[htb]
\caption{Attacks on ImageNet against \textbf{image recognition} models: VGG-16, Inception-v3 (Inc-v3), ResNet-50 (Res-50), MobileNet-v2 (MN-v2), ViT-Base (ViT-B/16), SwinTransformer-Tiny (Swin-T) and CLIP (Zero-Shot). Inception-v3 is utilized as the surrogate model (ACC denotes Accuracy).}
\setlength\tabcolsep{4pt}
\renewcommand\arraystretch{1.2}
\centering
\begin{tabular}{c|cc|cc|cc|cc|cc|cc|cc}
\hline
\multirow{2}{*}{Method} & \multicolumn{2}{c|}{VGG-16} & \multicolumn{2}{c|}{Inc-v3} & \multicolumn{2}{c|}{Res-50} & \multicolumn{2}{c|}{MN-v2} & \multicolumn{2}{c|}{ViT-B/16} & \multicolumn{2}{c|}{Swin-T} & \multicolumn{2}{c}{CLIP (Zero-Shot)} \\ \cline{2-15} 
                        & ACC~$\downarrow$          & NQ~$\downarrow$          & ACC~$\downarrow$          & NQ~$\downarrow$          & ACC~$\downarrow$          & NQ~$\downarrow$          & ACC~$\downarrow$         & NQ~$\downarrow$          & ACC~$\downarrow$        & NQ~$\downarrow$          & ACC~$\downarrow$          & NQ~$\downarrow$ 
                        & ACC~$\downarrow$         & NQ~$\downarrow$          \\ \hline

Clean                 &69.9\%               &  -           &66.9\%               & -            & 68.8\%               & -            & 75.4\%             &  -           & 78.2\%              &  -           & 78.9\%              &   -  & 67.3\%   &   -        \\ \hline
% DOP-Rd                  & 75.1\%              & 350            &  81.7\%             & 243            & 71.8\%              & 414            & 78.4\%             &  294           & 36.3\%             & 1781            & 44.7\%             &  1260           \\
DOP-LOA                 & 34.4\%               & 301            &  28.9\%             &  221           & 47.7\%              & 472            & 35.2\%             & 294            & 69.5\%             & 1152            & 64.1\%             & 927    & 40.6\%   &\textbf{347}            \\
DOP-DTA                 & \textbf{22.5\%}               & \textbf{191}            & \textbf{17.6\%}              &  \textbf{154}           &  \textbf{27.5\%}           &  \textbf{245}           &  \textbf{21.7\%}            &  \textbf{179}           & \textbf{61.4\%}             & \textbf{884}            & \textbf{54.3\%}             & \textbf{663}    & \textbf{39.1\%}   &360         \\ \hline
\end{tabular}
\label{tab:imagenet}
\end{table*}

\begin{figure*}[htb]
  \centering
  \includegraphics[width=0.95\linewidth]{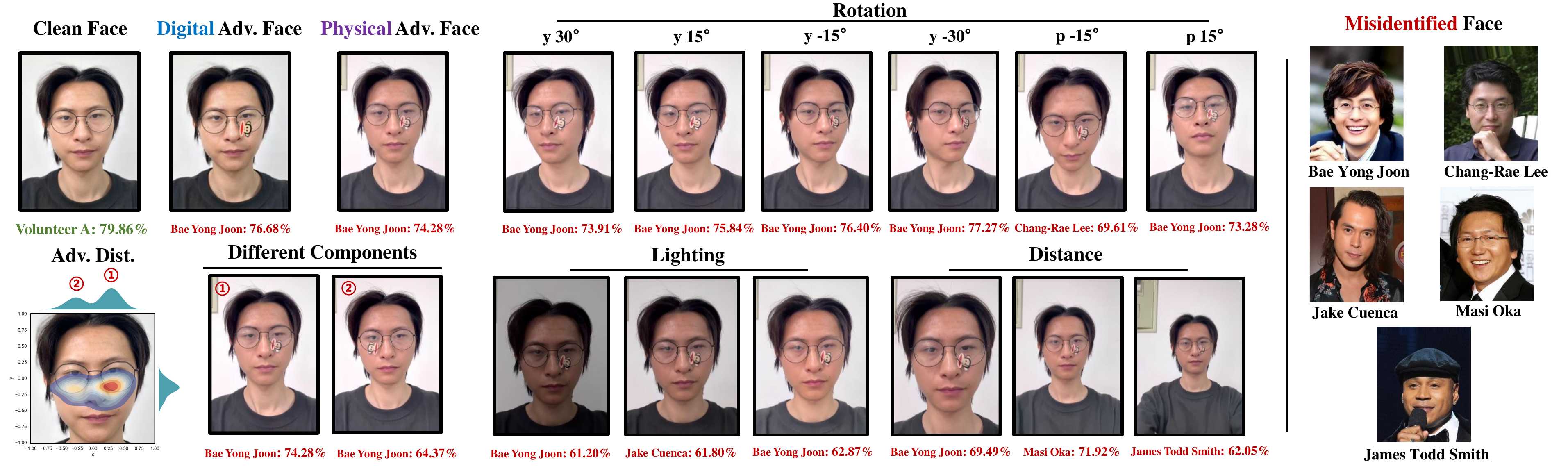}
   \caption{Qualitative results of \textbf{physical attacks} against face identification under different angles, distances, lighting, and locations.}
   \label{fig:phy} 
\end{figure*}

\begin{table}[htb]
\caption{Quantitative results of \textbf{physical attacks} against face identification under different angles, distances and lighting.}
\setlength\tabcolsep{3.5pt}
\renewcommand\arraystretch{1.2}
\centering
\begin{tabular}{c|c|c|c|c|c|c}
\hline
    & 0°    & yaw ±15° & yaw ±30° & pitch ±15° & distance  & lighting  \\ \hline
ASR & 85.42\% & 79.31\%    & 56.52\%    & 60.38\%      & 60.42\%    & 62.82\%    \\ \hline
\end{tabular}
\label{tab:quantitative}
\end{table}

\begin{table*}[htb]
\caption{Evaluation of \textbf{adversarial training} on face classification tasks using DOP-DMAT with FaceNet and ArcFace. Robustness is assessed based on the attack success rate (ASR) and the number of queries (NQ) under various attack methods. \emph{max} denotes the maximum queries are used.}
\setlength\tabcolsep{2.0pt}
\renewcommand\arraystretch{1.3}
\centering
\begin{tabular}{c|lcccccccclcccccccc}
\hline
\multirow{3}{*}{Method} & \multicolumn{9}{c|}{FaceNet}                                                                                                                                                                   & \multicolumn{9}{c}{ArcFace}                                                                                                                                                              \\ \cline{2-19} 
                        & \multicolumn{1}{c|}{\scriptsize Clean} & \multicolumn{2}{c|}{\scriptsize RHDE~\cite{wei2022adversarial}}       & \multicolumn{2}{c|}{\scriptsize GDPA~\cite{li2021generative}}     & \multicolumn{2}{c|}{\scriptsize DOP-LOA}     & \multicolumn{2}{c|}{\scriptsize DOP-DTA}     & \multicolumn{1}{c|}{\scriptsize Clean}  & \multicolumn{2}{c|}{\scriptsize RHDE~\cite{wei2022adversarial}}       & \multicolumn{2}{c|}{\scriptsize GDPA~\cite{li2021generative}}     & \multicolumn{2}{c|}{\scriptsize DOP-LOA}     & \multicolumn{2}{c}{\scriptsize DOP-DTA} \\ \cline{2-19} 
                        & \multicolumn{1}{l|}{\scriptsize{ASR~$\downarrow$}}  & \scriptsize{ASR~$\downarrow$} & \multicolumn{1}{c|}{\scriptsize{NQ~$\uparrow$}} & \scriptsize{ASR~$\downarrow$} & \multicolumn{1}{c|}{\scriptsize{NQ~$\uparrow$}} & \scriptsize{ASR~$\downarrow$} & \multicolumn{1}{c|}{\scriptsize{NQ~$\uparrow$}} & \scriptsize{ASR~$\downarrow$} & \multicolumn{1}{c|}{\scriptsize{NQ~$\uparrow$}} & \multicolumn{1}{l|}{\scriptsize{ASR~$\downarrow$}} & \scriptsize{ASR~$\downarrow$} & \multicolumn{1}{c|}{\scriptsize{NQ~$\uparrow$}} & \scriptsize{ASR~$\downarrow$} & \multicolumn{1}{c|}{\scriptsize{NQ~$\uparrow$}} & \scriptsize{ASR~$\downarrow$} & \multicolumn{1}{c|}{\scriptsize{NQ~$\uparrow$}} & \scriptsize{ASR~$\downarrow$}         & \scriptsize{NQ~$\downarrow$}         \\ \hline
\scriptsize Standard-Trained        & \multicolumn{1}{l|}{0.1\%}       & 42.2\%     & \multicolumn{1}{c|}{1945}    &53.0\%      & \multicolumn{1}{c|}{-}    &84.3\%      & \multicolumn{1}{c|}{110}    & 83.0\%     & \multicolumn{1}{c|}{142}    & \multicolumn{1}{l|}{1.1\%}      & 43.7\%     & \multicolumn{1}{c|}{1743}    &25.5\%      & \multicolumn{1}{c|}{-}    & 36.7\%     & \multicolumn{1}{c|}{895}    & 53.2\%             & 571             \\
Random-Aug              & \multicolumn{1}{l|}{0.0\%}      & 21.2\%      & \multicolumn{1}{c|}{4929}    &19.9\%      & \multicolumn{1}{c|}{-}    & 59.4\%     & \multicolumn{1}{c|}{447}    & 31.4\%     & \multicolumn{1}{c|}{1131}    & \multicolumn{1}{l|}{0.0\%}    &  12.8\%    & \multicolumn{1}{c|}{8539}    & 2.1\%      & \multicolumn{1}{c|}{-}    &7.8\%      & \multicolumn{1}{c|}{6560}    &2.2\%              &22450             \\ 

GDPA-AT~\cite{li2021generative}             & \multicolumn{1}{l|}{0.0\%}      & 11.4\%      & \multicolumn{1}{c|}{9725}    &2.5\%      & \multicolumn{1}{c|}{-}    & \textbf{0.8\%}     & \multicolumn{1}{c|}{\textbf{62018}}    & 15.4\%     & \multicolumn{1}{c|}{3096}    & \multicolumn{1}{l|}{0.0\%}    &  18.8\%    & \multicolumn{1}{c|}{5453}    & 1.0\%      & \multicolumn{1}{c|}{-}    &\textbf{0.0\%}      & \multicolumn{1}{c|}{\emph{max}}    &\textbf{0.9\%}              & \textbf{55475}             \\ 

\rowcolor{gray!30} DOP-DMAT              & \multicolumn{1}{l|}{0.0\%}       & \textbf{4.5\%}     & \multicolumn{1}{c|}{\textbf{26146}}    & \textbf{1.8\%}     & \multicolumn{1}{c|}{\textbf{-}}    &1.6\%     & \multicolumn{1}{c|}{30856}    &\textbf{6.0\%}      & \multicolumn{1}{c|}{\textbf{8334}}    & \multicolumn{1}{l|}{0.0\%}     & \textbf{7.8\%}     & \multicolumn{1}{c|}{\textbf{12817}}    &\textbf{0.1\%}     & \multicolumn{1}{c|}{\textbf{-}}    &\textbf{0.0\%}     & \multicolumn{1}{c|}{\emph{max}}    &1.4\%              &35950             \\ \hline
% DOP-DMAT(K=3)              & \multicolumn{1}{l|}{}      & \multicolumn{1}{c|}{}     &      & \multicolumn{1}{c|}{}    &      & \multicolumn{1}{c|}{}    &      & \multicolumn{1}{c|}{}    &      & \multicolumn{1}{c|}{}    & \multicolumn{1}{l|}{}      & \multicolumn{1}{c|}{}     &      & \multicolumn{1}{c|}{}    &      & \multicolumn{1}{c|}{}    &      & \multicolumn{1}{c|}{}    &              &             \\
% DOP-DMAT(K=5)              & \multicolumn{1}{l|}{}      & \multicolumn{1}{c|}{}     &      & \multicolumn{1}{c|}{}    &      & \multicolumn{1}{c|}{}    &      & \multicolumn{1}{c|}{}    &      & \multicolumn{1}{c|}{}    & \multicolumn{1}{l|}{}      & \multicolumn{1}{c|}{}     &      & \multicolumn{1}{c|}{}    &      & \multicolumn{1}{c|}{}    &      & \multicolumn{1}{c|}{}    &              &             \\ \hline

\end{tabular}
\label{tab:at}
\end{table*}

\begin{table}[htb]
\caption{Attack performance of RHDE on FaceNet using DOP-DMAT with different values of $K$.}
\setlength\tabcolsep{1.8pt}
\renewcommand\arraystretch{1.1}
\centering
\begin{tabular}{c|cc|cc|cc|cc|cc}
\hline
\multirow{2}{*}{} & \multicolumn{2}{c|}{$K$=1} & \multicolumn{2}{c|}{$K$=3} & \multicolumn{2}{c|}{$K$=5} & \multicolumn{2}{c|}{$K$=10} & \multicolumn{2}{c}{$K$=15} \\ \cline{2-11} 
                  & ASR         & NQ         & ASR        & NQ          & ASR        & NQ          & ASR        & NQ           & ASR        & NQ          \\ \hline
RHDE              & 12.9\%        & 8592       & 4.8\%        & 24280       & \textbf{4.5\%}        & \textbf{26146}       & 5.9\%        & 19698        & 5.6\%        & 21149       \\ \hline
\end{tabular}
\label{tab: ablation K AT} 
\end{table}

Besides, we also give physical attacks against face identification. We add a reference face image of a volunteer into the CelebA database, and then paste the sticker on the volunteer's face according to the results of DOP-DTA. Finally, we use a camera to shoot the volunteer's face. The captured manipulated face image is then used to perform face identification. To verify the robustness of our DOPatch, we test the attack performance under different angles (rotation), distances and lighting of the camera. The qualitative results are given in Fig. \ref{fig:phy}. Our DOP-DTA shows good attack ability under different physical changes, demonstrating its effectiveness in the real world. The quantitative results are given in Table \ref{tab:quantitative}, where we see that DOP-DTA achieves more than 50\% ASR under all the settings, a satisfactory physical attack performance.

\subsection{Attack Results on Image Recognition Task}
\begin{figure}[htb]
  \centering
  \includegraphics[width=0.99\linewidth]{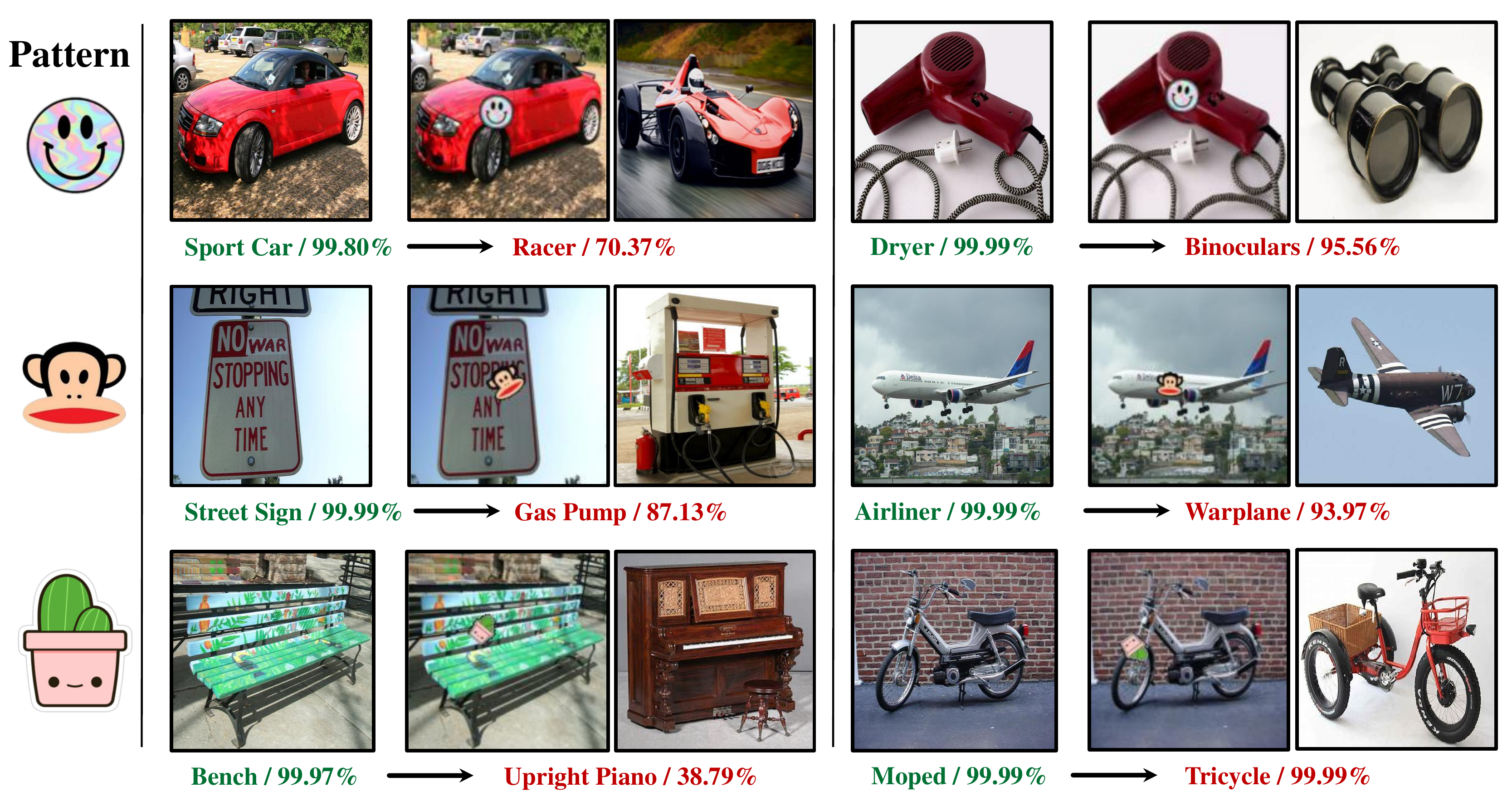}
   \caption{Examples of DOPatch against Inception-v3 classifier. Each group includes the original image, the attacked image, and the image corresponding to the predicted wrong class after the attacks.}
   \label{fig:vis2} 
\end{figure}

We further extend DOPatch to the image recognition task and conduct attack experiments on the ImageNet dataset. Specifically, we use classifiers pre-trained on ImageNet-1K as the target models, including CNN-based models VGG-16, Inception-v3, ResNet-50, and MobileNet-v2, as well as Transformer-based models ViT-Base, SwinTransformer-Tiny and Clip model, all these classifiers use open-source weights from torchvision\footnote[1]{\scriptsize\href{https://pytorch.org/vision/stable/index.html}{\textcolor{black}{https://pytorch.org/vision/stable/index.html}}} and timm\footnote[2]{\scriptsize\href{https://github.com/huggingface/pytorch-image-models}{\textcolor{black}{https://github.com/huggingface/pytorch-image-models}}}. We randomly selected 1000 images from the ImageNet-1K test set to conduct the attacks. In the attack setup, we use Inception-v3 as the surrogate model and keep the other parameter settings consistent with Sec.~\ref{sec:setting}. We evaluate the attack success rates and the number of queries of DOP-LOA and DOP-DTA on various classifiers, as shown in Tab.~\ref{tab:imagenet}. After applying the attacks, the classification performance of the models is significantly degraded, demonstrating the effectiveness of DOPatch. Some examples of the attacks are shown in Fig.~\ref{fig:vis2}. Furthermore, we observe that CNN-based models exhibit more noticeable performance degradation than Transformer-based models when facing DOPatch attacks, highlighting the robustness of Transformer-based models against patch perturbations.

\vspace{-3ex}
\subsection{Defense Results with Adversarial Training}
Table~\ref{tab:at} presents the performance of the proposed DOP-DMAT on the face classification task. We compare DOP-DMAT to a potential baseline, which involves using randomly positioned adversarial patch samples for data augmentation. Besides, we also compare with GDPA-AT \cite{li2021generative}, which utilizes a dynamic adversarial patch for both location and pattern to perform adversarial training. In our experimental setup, we first investigate the impact of $K$ in the inner maximization process to adversarial training. The results are given in Table \ref{tab: ablation K AT}. From the table, we can see that when $K=5$, DOP-DMAT achieves the best defensive performance, thus we fix $K\!\!=\!\!5$ to compare with other SOTA adversarial training methods in Table~\ref{tab:at}. The learning rate and lambda for the distribution mapping network are consistent with Sec.~\ref{sec:setting}. We set the learning rate to 0.001 for the face classification models and trained for 150 epochs. We conduct experiments on FaceNet and ArcFace models under the LFW dataset, evaluating the trained models with the proposed DOP-LOA, DOP-DTA and two location-aware attack methods: RHDE~\cite{wei2022adversarial} and  GDPA~\cite{li2021generative}. We have the following findings based on the results: DOP-DMAT perform best in improving the model's robustness against various location-aware patches compared to the random data augmentation and GDPA-AT that mainly utilize the individual locations to perform adversarial training, which can be attributed to DOP-DMAT's ability to capture the worst-case distribution of adversarial locations, thereby enhancing the model's robustness against these worst-case scenarios. Taking FaceNet as an example, after training by DOP-DMAT, the ASR of purely location-aware patches attack decreases from 42.2\% to 5.6\% (RHDE), 53.0\% to 1.8\% (GDPA),  84.3\% to 1.6\% (DOP-LOA) and 83.0\% to 6.0\% (DOP-DTA). Correspondingly, the query count also increases significantly. 
% Furthermore, we observed that although DOP-DMAT is specifically designed for adversarial training against fixed-pattern, the model's robustness can still generalize to attacks involving pattern optimizations, as shown in table~\ref{tab:at}. After DOP-DMAT training, the ASR of \cite{wei2022simultaneously} on FaceNet decreased from 90.3\% to xxx. Although it is higher than against purely position-aware methods, it still demonstrates a certain level of robustness. Further investigation into this phenomenon will be left for future work and discussion.

% Additionally, we observe that after random data augmentation, the performance of DOP-DTA deteriorates more significantly compared to DOP-Rd and DOP-LOA.

% or


\section{Conclusion}
In this paper, we have proposed a novel method for generating location-aware adversarial patches, called Distribution-Optimized Adversarial Patch (DOPatch). DOPatch optimized a multimodal distribution of adversarial locations instead of individual ones, which enabled efficient query-based black-box attacks via transfer prior, and robustness improvement via Distributional-Modeling Adversarial Training (DOP-DMAT). We have demonstrated the superiority and efficiency of DOPatch over existing methods on various face classification and recognition tasks. We have also provided insights into the distribution of adversarial locations and their relation to the model’s vulnerability. Our work has the potential to open up new possibilities for exploring and defending against location-aware adversarial patches in the physical world.


%\section{}

%\ifCLASSOPTIONcompsoc
%\else
%\fi


\ifCLASSOPTIONcaptionsoff
  \newpage
\fi



%\bibliographystyle{IEEEtran}

\bibliographystyle{splncs04}
\bibliography{egbib}

\begin{thebibliography}{10}
\providecommand{\url}[1]{\texttt{#1}}
\providecommand{\urlprefix}{URL }
\providecommand{\doi}[1]{https://doi.org/#1}

\bibitem{andriushchenko2020square}
Andriushchenko, M., Croce, F., Flammarion, N., Hein, M.: Square attack: a
  query-efficient black-box adversarial attack via random search. In: ECCV. pp.
  484--501 (2020)

\bibitem{athalye2018synthesizing}
Athalye, A., Engstrom, L., Ilyas, A., Kwok, K.: Synthesizing robust adversarial
  examples. In: International conference on machine learning. pp. 284--293.
  PMLR (2018)

\bibitem{blundell2015weight}
Blundell, C., Cornebise, J., Kavukcuoglu, K., Wierstra, D.: Weight uncertainty
  in neural network. In: International conference on machine learning. pp.
  1613--1622. PMLR (2015)

\bibitem{brown2017adversarial}
Brown, T.B., Man{\'e}, D., Roy, A., Abadi, M., Gilmer, J.: Adversarial patch.
  arXiv preprint arXiv:1712.09665  (2017)

\bibitem{carlini2017towards}
Carlini, N., Wagner, D.: Towards evaluating the robustness of neural networks.
  In: 2017 ieee symposium on security and privacy (sp). pp. 39--57. Ieee (2017)

\bibitem{chen2021camdar}
Chen, C., Huang, T.: Camdar-adv: generating adversarial patches on 3d object.
  International Journal of Intelligent Systems  \textbf{36}(3),  1441--1453
  (2021)

\bibitem{chen2018mobilefacenets}
Chen, S., Liu, Y., Gao, X., Han, Z.: Mobilefacenets: Efficient cnns for
  accurate real-time face verification on mobile devices. In: Biometric
  Recognition: 13th Chinese Conference, CCBR 2018, Urumqi, China, August 11-12,
  2018, Proceedings 13. pp. 428--438. Springer (2018)

\bibitem{chiang2020certified}
Chiang, P.y., Ni, R., Abdelkader, A., Zhu, C., Studer, C., Goldstein, T.:
  Certified defenses for adversarial patches. arXiv preprint arXiv:2003.06693
  (2020)

\bibitem{chindaudom2020adversarialqr}
Chindaudom, A., Siritanawan, P., Sumongkayothin, K., Kotani, K.: Adversarialqr:
  An adversarial patch in qr code format. In: 2020 Joint 9th International
  Conference on Informatics, Electronics \& Vision (ICIEV) and 2020 4th
  International Conference on Imaging, Vision \& Pattern Recognition (icIVPR).
  pp.~1--6. IEEE (2020)

\bibitem{danskin2012theory}
Danskin, J.M.: The theory of max-min and its application to weapons allocation
  problems, vol.~5. Springer Science \& Business Media (2012)

\bibitem{deng2019arcface}
Deng, J., Guo, J., Xue, N., Zafeiriou, S.: Arcface: Additive angular margin
  loss for deep face recognition. In: Proceedings of the IEEE/CVF conference on
  computer vision and pattern recognition. pp. 4690--4699 (2019)

\bibitem{dong2020adversarial}
Dong, Y., Deng, Z., Pang, T., Zhu, J., Su, H.: Adversarial distributional
  training for robust deep learning. Advances in Neural Information Processing
  Systems  \textbf{33},  8270--8283 (2020)

\bibitem{dong2018boosting}
Dong, Y., Liao, F., Pang, T., Su, H., Zhu, J., Hu, X., Li, J.: Boosting
  adversarial attacks with momentum. In: Proceedings of the IEEE conference on
  computer vision and pattern recognition. pp. 9185--9193 (2018)

\bibitem{eykholt2018robust}
Eykholt, K., Evtimov, I., Fernandes, E., Li, B., Rahmati, A., Xiao, C.,
  Prakash, A., Kohno, T., Song, D.: Robust physical-world attacks on deep
  learning visual classification. In: Proceedings of the IEEE conference on
  computer vision and pattern recognition. pp. 1625--1634 (2018)

\bibitem{goodfellow2014explaining}
Goodfellow, I.J., Shlens, J., Szegedy, C.: Explaining and harnessing
  adversarial examples. arXiv preprint arXiv:1412.6572  (2014)

\bibitem{guo2021meaningful}
Guo, Y., Wei, X., Wang, G., Zhang, B.: Meaningful adversarial stickers for face
  recognition in physical world. arXiv e-prints pp. arXiv--2104 (2021)

\bibitem{gurel2021knowledge}
G{\"u}rel, N.M., Qi, X., Rimanic, L., Zhang, C., Li, B.: Knowledge enhanced
  machine learning pipeline against diverse adversarial attacks. In:
  International Conference on Machine Learning. pp. 3976--3987 (2021)

\bibitem{hayes2018visible}
Hayes, J.: On visible adversarial perturbations \& digital watermarking. In:
  Proceedings of the IEEE Conference on Computer Vision and Pattern Recognition
  Workshops. pp. 1597--1604 (2018)

\bibitem{he2016deep}
He, K., Zhang, X., Ren, S., Sun, J.: Deep residual learning for image
  recognition. In: Proceedings of the IEEE conference on computer vision and
  pattern recognition. pp. 770--778 (2016)

\bibitem{huang2008labeled}
Huang, G.B., Mattar, M., Berg, T., Learned-Miller, E.: Labeled faces in the
  wild: A database forstudying face recognition in unconstrained environments.
  In: Workshop on faces in'Real-Life'Images: detection, alignment, and
  recognition (2008)

\bibitem{jaderberg2015spatial}
Jaderberg, M., Simonyan, K., Zisserman, A., et~al.: Spatial transformer
  networks. Advances in neural information processing systems  \textbf{28}
  (2015)

\bibitem{karmon2018lavan}
Karmon, D., Zoran, D., Goldberg, Y.: Lavan: Localized and visible adversarial
  noise. In: International Conference on Machine Learning. pp. 2507--2515. PMLR
  (2018)

\bibitem{kingma2013auto}
Kingma, D.P., Welling, M.: Auto-encoding variational bayes. arXiv preprint
  arXiv:1312.6114  (2013)

\bibitem{komkov2021advhat}
Komkov, S., Petiushko, A.: Advhat: Real-world adversarial attack on arcface
  face id system. In: 2020 25th International Conference on Pattern Recognition
  (ICPR). pp. 819--826. IEEE (2021)

\bibitem{kurakin2018adversarial}
Kurakin, A., Goodfellow, I.J., Bengio, S.: Adversarial examples in the physical
  world. In: Artificial intelligence safety and security, pp. 99--112. Chapman
  and Hall/CRC (2018)

\bibitem{lee2019physical}
Lee, M., Kolter, Z.: On physical adversarial patches for object detection.
  arXiv preprint arXiv:1906.11897  (2019)

\bibitem{li2021generative}
Li, X., Ji, S.: Generative dynamic patch attack. BMVC  (2021)

\bibitem{liu2019perceptual}
Liu, A., Liu, X., Fan, J., Ma, Y., Zhang, A., Xie, H., Tao, D.:
  Perceptual-sensitive gan for generating adversarial patches. In: Proceedings
  of the AAAI conference on artificial intelligence. vol.~33, pp. 1028--1035
  (2019)

\bibitem{liu2022segment}
Liu, J., Levine, A., Lau, C.P., Chellappa, R., Feizi, S.: Segment and complete:
  Defending object detectors against adversarial patch attacks with robust
  patch detection. In: Proceedings of the IEEE/CVF Conference on Computer
  Vision and Pattern Recognition. pp. 14973--14982 (2022)

\bibitem{liu2017sphereface}
Liu, W., Wen, Y., Yu, Z., Li, M., Raj, B., Song, L.: Sphereface: Deep
  hypersphere embedding for face recognition. In: Proceedings of the IEEE
  conference on computer vision and pattern recognition. pp. 212--220 (2017)

\bibitem{liu2018dpatch}
Liu, X., Yang, H., Liu, Z., Song, L., Li, H., Chen, Y.: Dpatch: An adversarial
  patch attack on object detectors. arXiv preprint arXiv:1806.02299  (2018)

\bibitem{liu2018large}
Liu, Z., Luo, P., Wang, X., Tang, X.: Large-scale celebfaces attributes
  (celeba) dataset. Retrieved August  \textbf{15}(2018), ~11 (2018)

\bibitem{madry2017towards}
Madry, A., Makelov, A., Schmidt, L., Tsipras, D., Vladu, A.: Towards deep
  learning models resistant to adversarial attacks. arXiv preprint
  arXiv:1706.06083  (2017)

\bibitem{metzen2021meta}
Metzen, J.H., Finnie, N., Hutmacher, R.: Meta adversarial training against
  universal patches. arXiv preprint arXiv:2101.11453  (2021)

\bibitem{mirsky2021ipatch}
Mirsky, Y.: Ipatch: A remote adversarial patch. arXiv preprint arXiv:2105.00113
   (2021)

\bibitem{naseer2019local}
Naseer, M., Khan, S., Porikli, F.: Local gradients smoothing: Defense against
  localized adversarial attacks. In: 2019 IEEE Winter Conference on
  Applications of Computer Vision (WACV). pp. 1300--1307. IEEE (2019)

\bibitem{nesti2022evaluating}
Nesti, F., Rossolini, G., Nair, S., Biondi, A., Buttazzo, G.: Evaluating the
  robustness of semantic segmentation for autonomous driving against real-world
  adversarial patch attacks. In: Proceedings of the IEEE/CVF Winter Conference
  on Applications of Computer Vision. pp. 2280--2289 (2022)

\bibitem{rao2020adversarial}
Rao, S., Stutz, D., Schiele, B.: Adversarial training against
  location-optimized adversarial patches. In: Computer Vision--ECCV 2020
  Workshops: Glasgow, UK, August 23--28, 2020, Proceedings, Part V 16. pp.
  429--448. Springer (2020)

\bibitem{schroff2015facenet}
Schroff, F., Kalenichenko, D., Philbin, J.: Facenet: A unified embedding for
  face recognition and clustering. In: Proceedings of the IEEE conference on
  computer vision and pattern recognition. pp. 815--823 (2015)

\bibitem{sharif2016accessorize}
Sharif, M., Bhagavatula, S., Bauer, L., Reiter, M.K.: Accessorize to a crime:
  Real and stealthy attacks on state-of-the-art face recognition. In:
  Proceedings of the 2016 acm sigsac conference on computer and communications
  security. pp. 1528--1540 (2016)

\bibitem{sharif2019general}
Sharif, M., Bhagavatula, S., Bauer, L., Reiter, M.K.: A general framework for
  adversarial examples with objectives. ACM Transactions on Privacy and
  Security (TOPS)  \textbf{22}(3),  1--30 (2019)

\bibitem{snoek2012practical}
Snoek, J., Larochelle, H., Adams, R.P.: Practical bayesian optimization of
  machine learning algorithms. Advances in neural information processing
  systems  \textbf{25} (2012)

\bibitem{sun2014deep}
Sun, Y., Wang, X., Tang, X.: Deep learning face representation from predicting
  10,000 classes. In: Proceedings of the IEEE conference on computer vision and
  pattern recognition. pp. 1891--1898 (2014)

\bibitem{szegedy2013intriguing}
Szegedy, C., Zaremba, W., Sutskever, I., Bruna, J., Erhan, D., Goodfellow, I.,
  Fergus, R.: Intriguing properties of neural networks. arXiv preprint
  arXiv:1312.6199  (2013)

\bibitem{taigman2014deepface}
Taigman, Y., Yang, M., Ranzato, M., Wolf, L.: Deepface: Closing the gap to
  human-level performance in face verification. In: Proceedings of the IEEE
  conference on computer vision and pattern recognition. pp. 1701--1708 (2014)

\bibitem{thys2019fooling}
Thys, S., Van~Ranst, W., Goedem{\'e}, T.: Fooling automated surveillance
  cameras: adversarial patches to attack person detection. In: Proceedings of
  the IEEE/CVF conference on computer vision and pattern recognition workshops.
  pp.~0--0 (2019)

\bibitem{wang2018cosface}
Wang, H., Wang, Y., Zhou, Z., Ji, X., Gong, D., Zhou, J., Li, Z., Liu, W.:
  Cosface: Large margin cosine loss for deep face recognition. In: Proceedings
  of the IEEE conference on computer vision and pattern recognition. pp.
  5265--5274 (2018)

\bibitem{wei2022adversarial}
Wei, X., Guo, Y., Yu, J.: Adversarial sticker: A stealthy attack method in the
  physical world. IEEE Transactions on Pattern Analysis and Machine
  Intelligence  \textbf{45}(3),  2711--2725 (2023)

\bibitem{wei2022simultaneously}
Wei, X., Guo, Y., Yu, J., Zhang, B.: Simultaneously optimizing perturbations
  and positions for black-box adversarial patch attacks. IEEE Transactions on
  Pattern Analysis and Machine Intelligence  \textbf{45}(7),  9041--9054 (2023)

\bibitem{xiang2021patchguard}
Xiang, C., Bhagoji, A.N., Sehwag, V., Mittal, P.: Patchguard: A provably robust
  defense against adversarial patches via small receptive fields and masking.
  In: USENIX Security Symposium. pp. 2237--2254 (2021)

\bibitem{yamanaka2020adversarial}
Yamanaka, K., Matsumoto, R., Takahashi, K., Fujii, T.: Adversarial patch
  attacks on monocular depth estimation networks. IEEE Access  \textbf{8},
  179094--179104 (2020)

\bibitem{yang2020patchattack}
Yang, C., Kortylewski, A., Xie, C., Cao, Y., Yuille, A.: Patchattack: A
  black-box texture-based attack with reinforcement learning. In: Computer
  Vision--ECCV 2020: 16th European Conference, Glasgow, UK, August 23--28,
  2020, Proceedings, Part XXVI. pp. 681--698. Springer (2020)

\bibitem{yang2023towards}
Yang, X., Liu, C., Xu, L., Wang, Y., Dong, Y., Chen, N., Su, H., Zhu, J.:
  Towards effective adversarial textured 3d meshes on physical face
  recognition. arXiv preprint arXiv:2303.15818  (2023)

\bibitem{zhu2017unpaired}
Zhu, J.Y., Park, T., Isola, P., Efros, A.A.: Unpaired image-to-image
  translation using cycle-consistent adversarial networks. In: Proceedings of
  the IEEE international conference on computer vision. pp. 2223--2232 (2017)

\bibitem{zolfi2021adversarial}
Zolfi, A., Avidan, S., Elovici, Y., Shabtai, A.: Adversarial mask: Real-world
  adversarial attack against face recognition models. arXiv preprint
  arXiv:2111.10759  (2021)

\end{thebibliography}

\end{document}

\appendices

\section{Pseudo-code for DOP-DTA and DOP-LOA}
To further help understand the optimization of DOP-DTA and DOP-LOA, we provide their pseudo-code here.

\begin{algorithm}[ht]
\footnotesize
\caption{\footnotesize Algorithms of DOP-DTA}\label{al:dopatch}
\KwIn{attack image $\mathbf{X}$, true label $y$, optimized distribution mapping network $F^*_\mathbf{W}$, treat model under black-box setting $f^\prime$, number of iterations $T$, components number $K$, Monte Carlo sample number (population size) $Q$, learning rate $\eta$, patch pattern $\mathbf{P}$.}
Initialize the distribution parameters: $\Psi^0=\{\omega^0_k , \boldsymbol{\mu}^0_k, \boldsymbol{\sigma}^0_k \}_{k=1}^K$, \text{where} 
$\{\omega^0_k , \boldsymbol{\mu}^0_k\}_{k=1}^K \leftarrow F^*_\mathbf{W}(\mathbf{X})$;\quad $\{\omega^0_k\}_{k=1}^K \leftarrow \frac{1}{K}$;

\For{$\mathrm{t}=1\;\mathbf{to}\; T$}{
\tcc{Monte Carlo sampling of variables}
Sample $\{\mathbf{r}_j\}_{j=1}^Q$ from $\mathcal{N}(\mathbf{0},\mathbf{I})$;

Sample $\{\boldsymbol{\Gamma}_j\}_{j=1}^Q$ from the multinomial distribution with probability $\omega^t_k$;

Calculate $\nabla_{\Psi^t}=\{\nabla_{\omega^t_k},\nabla_{\boldsymbol{\mu}^t_k},\nabla_{\boldsymbol{\sigma}^t_k}\}$ following Eq.~(16)  ;

\tcc{Update the distribution parameters}
$\Psi^{t+1} \leftarrow \Psi^{t}+\eta \cdot \nabla_{\Psi^t}$;

\tcc{Normalize component weights}
$\omega^{t+1}_k \leftarrow \omega^{t+1}_k/ {\textstyle \sum_{k=1}^{K}}\omega^{t+1}_k$;
}
\tcc{attack using optimized distribution}
Sample adversarial location $\theta^*$ from optimized distribution $\Psi^T=\{\omega^T_k , \boldsymbol{\mu}^T_k, \boldsymbol{\sigma}^T_k \}_{k=1}^K$ following Eq.~(4);

$\mathbf{X}_{\text{adv}} \leftarrow \mathcal{G}(\mathbf{X},\mathbf{P},\boldsymbol{\theta^*})$;

\KwOut{adversarial example $\mathbf{X}_{\text{adv}}$}

\end{algorithm}

\begin{algorithm}[ht]
\footnotesize
\caption{\footnotesize Algorithms of DOP-LOA}\label{al:dopatch}
\KwIn{attack image $\mathbf{X}$, true label $y$, optimized distribution mapping network $F^*_\mathbf{W}$, treat model under black-box setting $f^\prime$, Maximum number of queries $S$, number of observation sample $N$, patch pattern $\mathbf{P}$, objective function $\mathcal{L}_{\mathrm{adv}}(\cdot)$ }

\tcc{Obtain observation samples}
Obtain the distribution parameters: $\{ \boldsymbol{\mu}_k, \boldsymbol{\sigma}_k\}_{k=1}^K \leftarrow F^*_\mathbf{W}(\mathbf{X})$;

Choose the worst component: $\widetilde k \leftarrow \max_{k}  \mathcal{L}_{\mathrm{adv}}(\boldsymbol{\mu}_k)$;

$\mathcal{O}_N \leftarrow \{\boldsymbol{\theta}_i, \mathcal{L}_{\mathrm{adv}}(\boldsymbol{\theta}_i)\}_{i=1}^N ~ \text{where}~\boldsymbol{\theta}_i\in [\boldsymbol{\mu}_{\widetilde k} - 3\boldsymbol{\sigma}_{\widetilde k}, \boldsymbol{\mu}_{\widetilde k} + 3\boldsymbol{\sigma}_{\widetilde k}] $;

\For{$\mathrm{s}=0\;\mathbf{to}\; S-N$}{
\tcc{update the GP model}
$p(\boldsymbol{\theta}_{s+1}|\mathcal{O}_{N+s}) \sim \mathcal{N}(\tilde{\mu}_s(\boldsymbol{\theta}_{s+1}), \tilde\sigma^2_s(\boldsymbol{\theta}_{s+1}))$;

\tcc{Obtain the next point}
Select $\boldsymbol{\theta}_{s+1}$ following Eq.~(15);

$\mathbf{X}_{\text{adv}} \leftarrow \mathcal{G}(\mathbf{X},\mathbf{P},\boldsymbol{\theta_{s+1}})$;

\eIf{$f^\prime(\mathbf{X}) \ne y$}{break\;}{
$\mathcal{O}_{N+s} \leftarrow \mathcal{O}_{N+s} \cup \{\boldsymbol{\theta}_{s+1}, \mathcal{L}_{\mathrm{adv}}(\boldsymbol{\theta}_{s+1})\}$
}

}

\KwOut{adversarial example $\mathbf{X}_{\text{adv}}$}

\end{algorithm}

\section{Derivation of DOP-DTA}

In this section, we provide the derivation of Eq.~(16). 

Eq.~(16) provides the gradient computation formula for the distribution parameters w.r.t the optimization objective i.e., Eq.~(6). For the first term of Eq.~(6), which represents the expectation of the adversarial loss $\small \mathcal{L}_1=\mathbb{E}_{p(\mathbf{u},\boldsymbol{\Gamma}|\Psi)}\bigl [\mathcal{L}_{\text{adv}}\bigr]$, we follow the concept of \textbf{Natural Evolution Strategies (NES)~\cite{wierstra2014natural}}, approximate the gradients of the distribution parameters using \textbf{search gradients}. The search gradient is defined as the derivative of the log-likelihood of an arbitrary distribution w.r.t its distribution parameters. In DOPatch, we employ a multimodal Gaussian distribution, and the search gradient for the distribution parameters is given by:

\begin{equation}
\begin{split}
    &\nabla_{\boldsymbol{\omega}_k}\log p(\mathbf{u},\boldsymbol{\Gamma}|\Psi) = {\gamma_k}; \\
    &\nabla_{\boldsymbol{\mu}_k}\log p(\mathbf{u},\boldsymbol{\Gamma}|\Psi) =\frac{\mathbf{u}-\boldsymbol{\mu}_k}{\boldsymbol{\sigma}_k^2} \gamma_k=\frac{\boldsymbol{r}}{\boldsymbol{\sigma}_k} \gamma_k;\\ 
    &\nabla_{\boldsymbol{\sigma}_k}\log p(\mathbf{u},\boldsymbol{\Gamma}|\Psi) = \frac{(\mathbf{u}-\boldsymbol{\mu}_k)^2-\boldsymbol{\sigma}_k^2}{\boldsymbol{\sigma}_k^3} \gamma_k=\frac{\boldsymbol{r}^2-1}{\boldsymbol{\sigma}_k}\gamma_k,
\end{split}
\label{eq:a1}
\end{equation} 

where $\boldsymbol{r}$ follows the standard Gaussian distribution $\mathcal{N}(\textbf{0}, \mathbf{I})$. To alleviate the slow convergence issue of gradient ascent and approximate a more realistic gradient, NES further employs \textbf{natural gradients}~\cite{amari1998natural,amari1998natural_2} instead of ordinary stochastic gradients for updates. The natural gradient is defined as:

\begin{equation}
    \widetilde{\nabla}_{\omega_k, \boldsymbol{\mu}_k,\boldsymbol{\sigma}_k}\mathcal{L}_1 = \mathbf{F}^{-1}\nabla_{\omega_k, \boldsymbol{\mu}_k,\boldsymbol{\sigma}_k}\mathcal{L}_1,
    \label{eq:a2}
\end{equation}

where $\mathbf{F}$ is the \textbf{Fisher information matrix} as:

\begin{equation}
\begin{split}
    \mathbf{F} = \mathbb{E}_{p(\mathbf{u},\boldsymbol{\Gamma}|\Psi)} &\big[\nabla_{\omega_k, \boldsymbol{\mu}_k,\boldsymbol{\sigma}_k}\log p(\mathbf{u},\boldsymbol{\Gamma}|\Psi) \\ &\cdot \nabla_{\omega_k, \boldsymbol{\mu}_k,\boldsymbol{\sigma}_k}\log p(\mathbf{u},\boldsymbol{\Gamma}|\Psi)^\top\big],
\end{split}
\label{eq:a3}
\end{equation}

By integrating Eq.~\eqref{eq:a1}, Eq.~\eqref{eq:a2} and Eq.~\eqref{eq:a3}, we can derive the modified gradient of the distribution parameters:

\begin{equation}
\begin{split}
    &\widetilde\nabla_{\boldsymbol{\omega}_k}\log p(\mathbf{u},\boldsymbol{\Gamma}|\Psi)=\mathbf{F}_{\omega_k}^{-1}{\gamma_k} = \frac{\gamma_k}{\omega_k}; \\
    &\widetilde\nabla_{\boldsymbol{\mu}_k}\log p(\mathbf{u},\boldsymbol{\Gamma}|\Psi) =\mathbf{F}_{\boldsymbol{\mu}_k}^{-1} \frac{\boldsymbol{r}}{\boldsymbol{\sigma}_k} \gamma_k = \frac{\boldsymbol{\sigma}_k\mathbf{r}}{\omega_k}\gamma_k;\\
    &\widetilde\nabla_{\boldsymbol{\sigma}_k}\log p(\mathbf{u},\boldsymbol{\Gamma}|\Psi)= \mathbf{F}_{\boldsymbol{\sigma}_k}^{-1}\frac{\boldsymbol{r}^2-1}{\boldsymbol{\sigma}_k}\gamma_k = \frac{\boldsymbol{\sigma}_k(\mathbf{r}-1)^2}{2\omega_k}\gamma_k.
\end{split}
\label{eq:a4}
\end{equation}

By incorporating Eq.~\eqref{eq:a4} into Eq.~(6), we complete the first part derivation, the gradient of the adversarial loss expectation w.r.t the distribution parameters is given by:

\begin{equation}
\begin{split}
    &\nabla_{\omega_k}\mathcal{L}_1 = \mathbb{E}_{\mathcal{N}(\mathbf{r}|\mathbf{0},\mathbf{I})}\big [\mathcal{L}_\text{cls} \cdot \frac{\gamma_k}{\omega _k} \big ]; \\
    &\nabla_{\boldsymbol{\mu}_k}\mathcal{L}_1 = \mathbb{E}_{\mathcal{N}(\mathbf{r}|\mathbf{0},\mathbf{I})}\big [\mathcal{L}_\text{cls} \cdot \frac{\boldsymbol{\sigma}_k \mathbf{r} }{\omega _k} \gamma _k\big]; \\ &\nabla_{\boldsymbol{\sigma}_k}\mathcal{L}_1 = \mathbb{E}_{\mathcal{N}(\mathbf{r}|\mathbf{0},\mathbf{I})}\big [\mathcal{L}_\text{cls} \cdot \frac{\boldsymbol{\sigma}_k (\mathbf{r}^2-1) }{2\omega_k}\gamma _k \big]. 
\end{split}
\label{eq:a5}
\end{equation}

For the second term of Eq.~(6), i.e., the expectation of entropy regularization loss $\small \mathcal{L}_2\!=\!\mathbb{E}_{p(\mathbf{u},\boldsymbol{\Gamma}|\Psi)}\bigl [ - \log p(\tanh(\mathbf{u}/\tau)) \bigr].$ We use the transformation of random variables to rewrite it as $\small \mathcal{L}_2\!=\!\mathbb{E}_{p(\mathbf{u},\boldsymbol{\Gamma}|\Psi)}\bigl [-\log p(\tanh((\prod_{k=1}^{K}\boldsymbol{\mu}^{\gamma_k}_k+{ \prod_{k=1}^{K}}\boldsymbol{\sigma}^{\gamma_k}_k\cdot\mathbf{r})/ \tau) \bigr].$ The log density of the distribution can be analytically calculated as follows. Note that the dimensions of the random variables are independent of each other, we consider here the case of one dimension. For the random variable $r$, its probability density is $p(r)=\frac{1}{\sqrt{2\pi} } \exp (-\frac{r^2}{2} )$. The probability density of $u$ is $p(u)=\prod_{k=1}^{K} \omega_k^{\gamma_k}(\frac{1}{\sqrt{2\pi}\sigma_k } \exp (-\frac{r^2}{2} ))^{\gamma_k}$. For the location parameter $\theta= \tanh(u/\tau)$, its inversion is $u=\tanh^{-1}(\theta)\cdot\tau=\frac{\tau}{2}\log(\frac{1+\theta}{1-\theta})$, The derivative of $u$ w.r.t. $\theta$ is $\frac{\mathrm{d} u}{\mathrm{d} \theta} = \frac{\tau}{\theta^2-1}$. By applying the transformation of variable approach, we can derive the probability density of $\theta$ as:

\begin{equation}
\begin{split}
    p(v)=&\prod_{k=1}^{K}(\omega_k\frac{1}{\sqrt{2\pi}\sigma_k } \exp (-\frac{r^2}{2} ))^{\gamma_k} \cdot \\ &\frac{\tau}{\tanh(\prod_{k=1}^{K}\frac{\mu^{\gamma_k}_k}{\tau}+{ \prod_{k=1}^{K}}\frac{\sigma^{\gamma_k}_k\cdot r}{\tau})^2-1}.
\end{split}
\end{equation}

Through the aforementioned transformation, we can derive the negative log-likelihood of the random variable $\theta$ as follow, which is also the formula for calculating the entropy regularization loss in Eq.~(10):

\begin{equation}
\begin{split}
    -\log p(v)=& \sum_{k=1}^{K}\gamma_k\Big[-\omega_k+ \frac{r^2}{2}+\frac{\log(2\pi)}{2}+\log \sigma_k \\ & +\log(\tanh(\frac{\mu_k+\sigma_k r}{\tau})^2-1)-\log \tau \Big].
\end{split}
\end{equation}

Sum over all dimensions, we can simply calculate the gradients of $\mathcal{L}_2$ w.r.t. $\omega_k$, $\boldsymbol{\mu}_k$, and $\boldsymbol{\sigma}_k$ as:

\begin{equation}
\begin{split}
    &\nabla_{\omega_k}\mathcal{H} = \mathbb{E}_{\mathcal{N}(\mathbf{r}|\mathbf{0},\mathbf{I})} [-\gamma_k];\\ &\nabla_{\boldsymbol{\mu}_k}\mathcal{H} = \mathbb{E}_{\mathcal{N}(\mathbf{r}|\mathbf{0},\mathbf{I})}\big [-2\gamma_k \tanh(\frac{\boldsymbol{\mu}_k+\boldsymbol{\sigma}_k \mathbf{r}}{\tau})\cdot\frac{1}{\tau}\big];\\ 
    &\nabla_{\boldsymbol{\sigma}_k}\mathcal{H} = \mathbb{E}_{\mathcal{N}(\mathbf{r}|\mathbf{0},\mathbf{I})}\big [\gamma_k \frac{1-2\tanh(\frac{\boldsymbol{\mu}_k+\boldsymbol{\sigma}_k\mathbf{r}}{\tau})\cdot\frac{\boldsymbol{\sigma}_k \mathbf{r}}{\tau}}{\boldsymbol{\sigma}_k} \big].
\end{split}
\label{eq:a8}
\end{equation}

Finally, we can obtain the gradient in Eq.~(16) by combining Eq.\eqref{eq:a5} and Eq.\eqref{eq:a8}.

\bibliographystyle{splncs04}
\bibliography{egbib}